\documentclass[11pt, web]{article}

\usepackage[margin=16mm]{geometry}
\usepackage{setspace}

\usepackage{times} 

\usepackage{amsmath, amssymb, amsthm}
\usepackage{mathtools} 

\usepackage{booktabs}       
\usepackage{array}          
\usepackage{multirow}       
\usepackage{diagbox}        

\usepackage{graphicx}       
\usepackage{caption}        
\usepackage{subcaption}     

\usepackage[ruled,vlined]{algorithm2e} 

\usepackage[english]{babel} 
\usepackage{lineno}         

\usepackage{xcolor}         
\usepackage[colorlinks=true,
            linkcolor=blue,
            citecolor=blue,
            urlcolor=red,
            pdfencoding=auto]{hyperref}

\newcommand{\cvect}{\mathbf{c}}

\newcommand{\svect}{\mathbf{s}}

\newcommand{\yvect}{\mathbf{y}}
\newcommand{\zvect}{\mathbf{z}}

\usepackage{authblk} 

\makeatletter
\renewcommand\AB@affilsepx{, \protect\Affilfont}
\makeatother
\usepackage{tabularx}
\providecommand{\keywords}[1]
{
  \small	
  \textbf{\textit{Keywords---}} #1
}

\begin{document}

\title{\textbf{SCS-SupCon: Sigmoid-based Common and Style Supervised Contrastive Learning with Adaptive Decision Boundaries}}
\author[1]{Bin Wang}
\author[1,2]{Fadi Dornaika\thanks{Corresponding author}}
\affil[1]{University of the Basque Country}
\affil[2]{IKERBASQUE, Basque Foundation for Science}

\affil[ ]{
\small\texttt{     \hspace{4cm}    bwang001@ikasle.ehu.eus, fadi.dornaika@ehu.eus}}
\date{}
\maketitle

\begin{abstract}
Image classification is hindered by subtle inter-class differences and substantial intra-class variations, which limit the effectiveness of existing contrastive learning methods. Supervised contrastive approaches based on the InfoNCE loss suffer from negative-sample dilution and lack adaptive decision boundaries, thereby reducing discriminative power in fine-grained recognition tasks. To address these limitations, we propose Sigmoid-based Common and Style Supervised Contrastive Learning (SCS-SupCon). Our framework introduces a sigmoid-based pairwise contrastive loss with learnable temperature and bias parameters to enable adaptive decision boundaries. This formulation emphasizes hard negatives, mitigates negative-sample dilution, and more effectively exploits supervision. In addition, an explicit style-distance constraint further disentangles style and content representations, leading to more robust feature learning. Comprehensive experiments on six benchmark datasets, including CUB200-2011 and Stanford Dogs, demonstrate that SCS-SupCon achieves state-of-the-art performance across both CNN and Transformer backbones. On CIFAR-100 with ResNet-50, SCS-SupCon improves top-1 accuracy over SupCon by approximately 3.9 percentage points and over CS-SupCon by approximately 1.7 points under five-fold cross-validation. On fine-grained datasets, it outperforms CS-SupCon by 0.4--3.0 points. Extensive ablation studies and statistical analyses further confirm the robustness and generalization of the proposed framework, with Friedman tests and Nemenyi post-hoc evaluations validating the stability of the observed improvements.

\end{abstract}

\keywords{ Supervised contrastive learning; Fine-grained classification;  Negative sample dilution;  Feature disentanglement; Decision boundary adjustment }
 \hspace{10pt}

\section{Introduction}

Fine-grained image classification has attracted considerable attention due to its broad applicability and significant real-world value, such as species identification~\cite{wah2011caltech}, medical image diagnosis~\cite{wei2021fine}, and industrial defect inspection~\cite{ma2024surface}. However, fine-grained classification remains inherently challenging, as subtle inter-class differences and substantial intra-class variations make conventional classification methods insufficient for achieving high accuracy and robust generalization~\cite{wei2021fine,lim2025fssf}. Therefore, effectively learning discriminative features that capture subtle distinctions among visually similar categories is a critical task in computer vision research~\cite{lim2025fssf,zhao2017survey}.

Current mainstream approaches addressing these challenges can be broadly categorized into four types: InfoNCE-based supervised contrastive learning methods~\cite{oord2018representation} (e.g., SupCon~\cite{khosla2020supervised}, SelfCon~\cite{baeSelfContrastiveLearningSingleviewed2022}, CS-SupCon~\cite{dornaika2025deep}, TimeSCL~\cite{huang2025noise}, FNCL~\cite{huynh2022boosting}), negative-free methods (e.g., BYOL~\cite{Grill2020}, SimSiam~\cite{Chen2021}), clustering-based methods (e.g., PCL~\cite{li2021prototypical}, SwAV~\cite{Caron2020}, CSTCN~\cite{wang2024learning}, Circle Loss~\cite{gui2025cross}), and hybrid contrastive approaches (e.g., PaCo~\cite{cui2021parametric}, DACL~\cite{xu2025few}, CSA-RSIC~\cite{cheng2025csa}). Among these, InfoNCE-based supervised contrastive methods are notably susceptible to negative-sample dilution, since each anchor is indiscriminately compared against numerous negatives, diluting the gradients from critical hard negatives~\cite{khosla2020supervised,wang2021understanding}. Negative-free methods circumvent dilution but often suffer from feature collapse issues due to the absence of explicit negative supervision~\cite{zbontar2021barlow}. Clustering-based methods partially alleviate dilution through semantic grouping but typically lack explicit adaptive decision boundaries, restricting precise discrimination~\cite{li2021prototypical,Caron2020}. Hybrid methods combine multiple strategies yet frequently introduce excessive complexity and sensitivity to hyperparameters, limiting practical usability~\cite{cui2021parametric,xu2025few}.

\begin{figure}[htbp]
    \centering
    \includegraphics[width=1.05\textwidth]{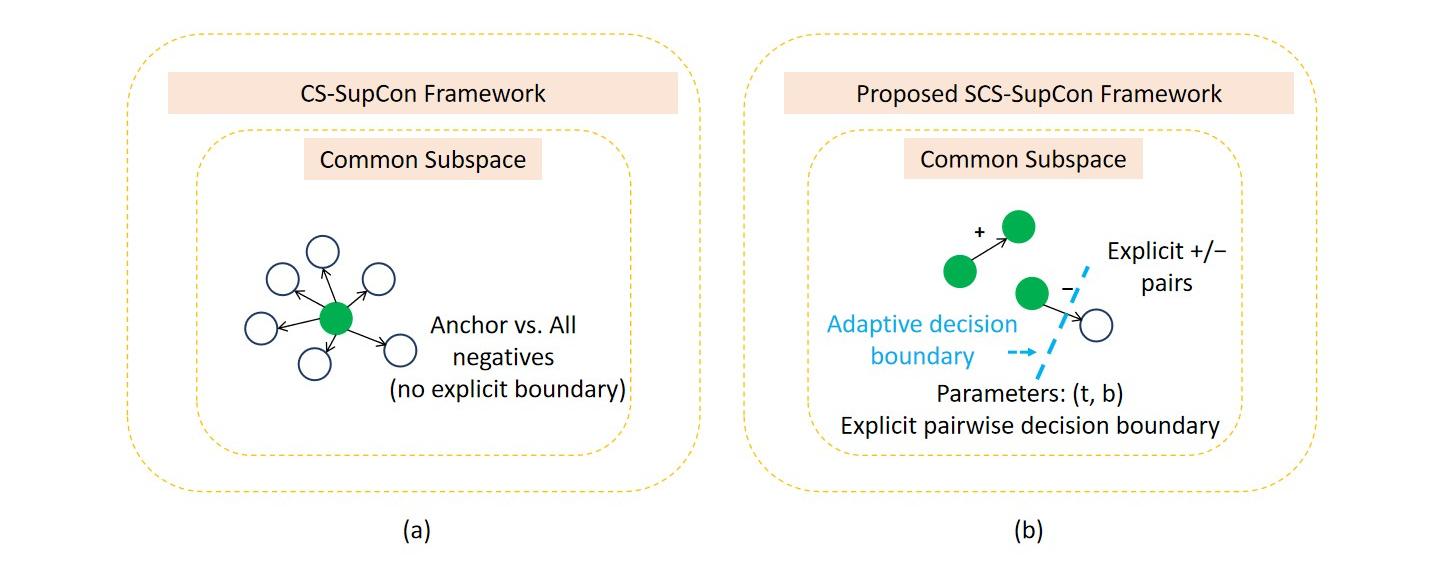}
    \caption{
        Motivation and core innovations of our proposed SCS-SupCon method.
        (a) Existing CS-SupCon employs an InfoNCE-based contrastive loss, causing dilution of negative-sample information by simultaneously comparing numerous negatives.
        (b) Our proposed SCS-SupCon explicitly introduces a sigmoid-based pairwise contrastive loss with learnable parameters (temperature $t$ and bias $b$), adaptively adjusting decision boundaries to effectively distinguish subtle differences between negative and positive pairs.
    }
    \label{fig:SCS_intro}
\end{figure}

This paper specifically targets improvements in InfoNCE-based supervised contrastive learning. Although InfoNCE-based methods have shown powerful representation capabilities, their performance in fine-grained tasks remains severely limited by negative-sample dilution and the absence of explicit decision boundary mechanisms (see Figure~\ref{fig:SCS_intro}(a)). Existing methods typically treat negative samples equally, diluting informative gradients from hard negatives. Additionally, without explicit decision boundary mechanisms, current methods cannot adaptively define differentiation thresholds between positive and negative pairs. Recently, an explicit style-distance constraint has been introduced in CS-SupCon~\cite{dornaika2025deep} to enhance feature disentanglement, which demonstrated promising potential but still relies fundamentally on the InfoNCE framework.

Consequently, current supervised contrastive learning methods urgently require solutions to the following key challenges:

\textbf{Negative-Sample Dilution Problem:}  
Existing supervised contrastive methods employing InfoNCE-based losses indiscriminately compare anchor samples against multiple negatives simultaneously, diluting crucial negative-sample information and impairing discriminative capability (Figure~\ref{fig:SCS_intro}(a)).

\textbf{Lack of Explicit Decision Boundary Adjustment Mechanisms:}  
Current supervised contrastive losses implicitly handle intra-class compactness without explicit, adaptive mechanisms to control decision boundaries, limiting precise differentiation between positive and negative pairs.

To address these critical issues, we propose a novel supervised contrastive learning framework termed SCS-SupCon (Sigmoid-based Common and Style Supervised Contrastive Learning), as depicted in Figure~\ref{fig:SCS_intro}(b). Our proposed SCS-SupCon explicitly introduces a sigmoid-based pairwise contrastive loss with adaptive parameters (temperature and bias), enabling precise and flexible control over decision boundaries between positives and negatives, significantly mitigating negative-sample dilution. Furthermore, we inherit and extend the explicit style-distance constraint from CS-SupCon, effectively disentangling common and style features to substantially enhance discriminative power and generalization capability for fine-grained classification tasks.

We conducted comprehensive experiments on six benchmark image datasets using various backbone encoders. The results summarized in Tables~\ref{tab:results} and~\ref{tab:my-table} clearly indicate that our proposed SCS-SupCon consistently achieves state-of-the-art performance, surpassing classical contrastive methods (SupCon, SimCLR, BYOL), recent supervised approaches (SelfCon, CS-SupCon), clustering-based methods (PCL, CSTCN), and hybrid methods (PaCo). Our sigmoid-based contrastive loss effectively mitigates negative-sample dilution prevalent in conventional methods, resulting in substantial accuracy improvements. These improvements are particularly pronounced on fine-grained datasets such as CUB200-2011 and Stanford Dogs and are consistently validated across both CNN and Transformer backbones. Rigorous ablation studies and statistical analyses further confirm our method's effectiveness and robustness. Moreover, statistical significance tests (Friedman test~\cite{friedman1937use} with Nemenyi post-hoc analysis~\cite{demvsar2006statistical}) verify the stability and reliability of our improvements across multiple datasets.

\textcolor{black}{Beyond highlighting the empirical limitations of existing InfoNCE-based approaches, it is important to clarify why we adopt a sigmoid-based contrastive loss formulation with learnable temperature and bias rather than other boundary-adaptive alternatives (e.g., margin-based or softmax-adaptive losses). In classical margin-based SCL, the decision boundary is typically adjusted in the logit space of a softmax classifier, which implicitly couples the boundary control with the classification head and relies on a fixed set of class prototypes. In contrast, our sigmoid-based pairwise loss directly operates on pairwise similarities in the embedding space and exposes the decision boundary through two interpretable and learnable parameters $(t',b)$ that are learned jointly with the encoder and projection head. This design allows SCS-SupCon to reshape the effective margin between positive and negative pairs in a data-driven way, while remaining agnostic to the specific classifier architecture and preserving the simplicity of the underlying backbone.}

The main contributions of this paper are summarized as follows:

\begin{itemize}
    \item \textbf{Sigmoid-based Contrastive Loss Design:}  
    We introduce a novel sigmoid-based pairwise contrastive loss to supervised contrastive learning, effectively mitigating negative-sample dilution and enabling pair-specific differentiation, a significant advantage over traditional InfoNCE losses.

    \item \textbf{Adaptive Decision Boundary Adjustment Mechanism:}  
    By incorporating learnable temperature and bias parameters, our SCS-SupCon explicitly establishes adaptive decision boundaries, significantly enhancing discrimination of subtle differences between positive and negative pairs.

    \item \textbf{Explicit Style Distance Constraint:} 
    We further enhance the explicit style-distance constraint by seamlessly integrating it into our sigmoid-based contrastive framework, effectively disentangling style features from common features and significantly improving the robustness and discriminative capability of learned representations.

    \item \textbf{Significant Generalization Improvement Across Datasets and Architectures:}  
    Extensive experiments demonstrate that SCS-SupCon consistently and significantly outperforms existing supervised contrastive methods across diverse datasets and architectures, excelling notably in fine-grained classification tasks.
\end{itemize}

\textcolor{black}{The remainder of this paper is organized as follows: Section~\ref{sec2} reviews related contrastive learning methods and recent advances; Section~\ref{sec3} introduces the fundamentals of CS-SupCon; Section~\ref{sec4} details the proposed SCS-SupCon method; Section~\ref{sec5} presents comprehensive experimental results and analysis; Section~\ref{sec6} discusses the implications, limitations, and potential extensions of our approach; and finally, Section~\ref{sec7} concludes the paper.}

\section{Related Work}
\label{sec2}

We review supervised contrastive learning methods from the perspective of loss function design, specifically addressing two key challenges: \textit{negative-sample dilution} (where excessive negative examples weaken gradient signals) and the lack of \textit{explicit adaptive decision boundaries} (clear margins separating positives from negatives).

\subsection{InfoNCE-Based Contrastive Methods}

InfoNCE loss~\cite{oord2018representation} underpins influential methods like SimCLR~\cite{Chen2020} and MoCo~\cite{he2020momentum}, which maximize mutual information between positive pairs against negatives. However, due to indiscriminate treatment of negative pairs, these methods suffer from severe negative-sample dilution~\cite{wang2021understanding}. Variants including SupCon~\cite{khosla2020supervised}, SelfCon~\cite{baeSelfContrastiveLearningSingleviewed2022}, and CS-SupCon~\cite{dornaika2025deep} mitigate dilution through supervision or confidence weighting but inherit the fundamental drawbacks of fixed negative sample weighting.

Recent methods like DCL~\cite{yeh2021decoupled} and FNCL~\cite{huynh2022boosting} decouple positives and negatives for better gradient control. TimeSCL~\cite{huang2025noise} proposes temporally adaptive weighting for time-series data. Yet, all these approaches rely on predefined contrastive pairs, failing to establish learnable decision boundaries.

\subsection{Negative-Free Contrastive Methods}

To circumvent dilution entirely, methods such as BYOL~\cite{Grill2020} and SimSiam~\cite{Chen2021} adopt architectural asymmetry without negative samples. Nonetheless, the lack of explicit negatives restricts their discriminative power, posing significant limitations in fine-grained classification scenarios~\cite{zbontar2021barlow,bardes2022vicreg}.

\subsection{Clustering-Based Contrastive Methods}

PCL~\cite{li2021prototypical} and SwAV~\cite{Caron2020} introduce prototypes or clustering mechanisms to implicitly define negatives. CSTCN~\cite{wang2024learning} further integrates clustering cues into spatio-temporal learning frameworks. Despite these advances, clustering-based approaches offer limited explicit control over decision boundaries. Meanwhile, Circle Loss~\cite{gui2025cross}, adapted from metric learning, introduces angular margin adjustments but requires prototype labels, complicating model training.

\subsection{Hybrid and Adversarial Contrastive Losses}

Recent hybrid methods like PaCo~\cite{cui2021parametric}, CoMatch~\cite{li2021comatch}, and CSA-RSIC~\cite{cheng2025csa} blend contrastive mechanisms with prototypes or consistency-based regularization. DACL~\cite{xu2025few} employs dynamic adversarial margins to improve boundary clarity. Although effective, these hybrid methods often depend on additional architectural complexity or intricate training schedules. \textcolor{black}{We also note an important difference between prototype-based clustering methods and our explicit boundary formulation in the context of class-imbalanced and fine-grained data. Prototype-based approaches typically summarise each class by a small number of centroids; in long-tailed scenarios, these centroids can be dominated by frequent classes, which may limit their ability to capture the diversity of rare categories. Our proposed SCS-SupCon loss, by contrast, defines supervision at the level of individual pairwise similarities and modulates their influence through two interpretable and learnable parameters, allowing the model to preserve meaningful margins for under-represented but visually subtle classes.}

\subsection{Metric-Based and Adaptive Boundary Methods}

\textcolor{black}{Beyond the categories discussed above, a growing body of work has explored metric-based and adaptive-boundary strategies that adjust decision surfaces or loss landscapes in a data-driven manner. In applied domains such as medical imaging, structural health monitoring, and time-series forecasting, bio-inspired or meta-heuristic optimizers have been used to tune classifier parameters or feature-selection pipelines, often improving robustness under domain-specific constraints~\cite{elkenawy2024greylag,khodadadi2024data,karim2023novel,elshewey2024eeg,ebrahim2025liver}. These approaches typically operate at the classifier level, for example by adapting margins in the logit space or by optimizing decision thresholds through population-based search. In contrast, our proposed SCS-SupCon method acts directly on the deep features and integrates pair-similarity boundaries into the supervised contrastive loss: the learnable temperature and bias parameters modulate the pairwise similarity distribution in the embedding space, offering a compact and architecture-agnostic mechanism for controlling the effective margin between positive and negative pairs. }

\textcolor{black}{Recent advances have also explored diffusion-based contrastive frameworks and hierarchical transformer-based supervised contrastive methods for fine-grained recognition~\cite{ma2025cocova,jing2025interest}, which further enrich the landscape of adaptive contrastive learning. In particular, diffusion-augmented contrastive models use a generative diffusion process to produce informative views while a contrastive objective is optimised in the latent space, so they can be viewed as diffusion-based, adaptive-boundary contrastive methods rather than plain supervised SCL baselines. These techniques are largely complementary to our proposal and could be combined with SCS-SupCon in future work to investigate their joint benefits on larger and more diverse benchmarks.}

\subsection{Our Proposed Sigmoid-Based Contrastive Loss}

To overcome these limitations, we propose SCS-SupCon, a novel sigmoid-based pairwise contrastive loss featuring learnable temperature and bias parameters. Our design explicitly addresses negative-sample dilution by adaptively weighting negative examples and establishes clear, learnable decision boundaries. Compared to InfoNCE-based and clustering losses, SCS-SupCon significantly enhances discriminative capability without increasing complexity. Unlike adversarial methods like DACL or handcrafted temporal schemes like TimeSCL, SCS-SupCon provides an elegant and lightweight solution suitable for fine-grained classification.

\section{Preliminaries}
\label{sec3}

In this section, we concisely revisit the essential components and training strategy of the recently proposed CS-SupCon framework~\cite{dornaika2025deep}, which provides the foundational basis and motivation for our proposed SCS-SupCon approach.

\subsection{Overview of CS-SupCon}

Analogous to the conventional supervised contrastive learning (SupCon) framework~\cite{khosla2020supervised}, the CS-SupCon method primarily comprises three modules: (i) an augmentation module, producing multiple augmented representations for each input image; (ii) an encoder responsible for extracting deep image features; and (iii) a projection head that maps the deep features into normalized embeddings suited for metric learning.

The core novelty of CS-SupCon lies in explicitly decomposing the embedding space into two separate subspaces: a \emph{common subspace} $\cvect \in \mathbb{R}^{D_c}$ encoding class-relevant characteristics, and a \emph{style subspace} $\svect \in \mathbb{R}^{D_s}$ capturing class-irrelevant variations. Formally, the final embedding vector $\zvect$ is defined as follows:
\begin{equation}
    \zvect = [\cvect; \svect], \quad D_p = D_c + D_s.
\end{equation}

\begin{figure}[!htbp]
    \centering
    \includegraphics[width=\textwidth]{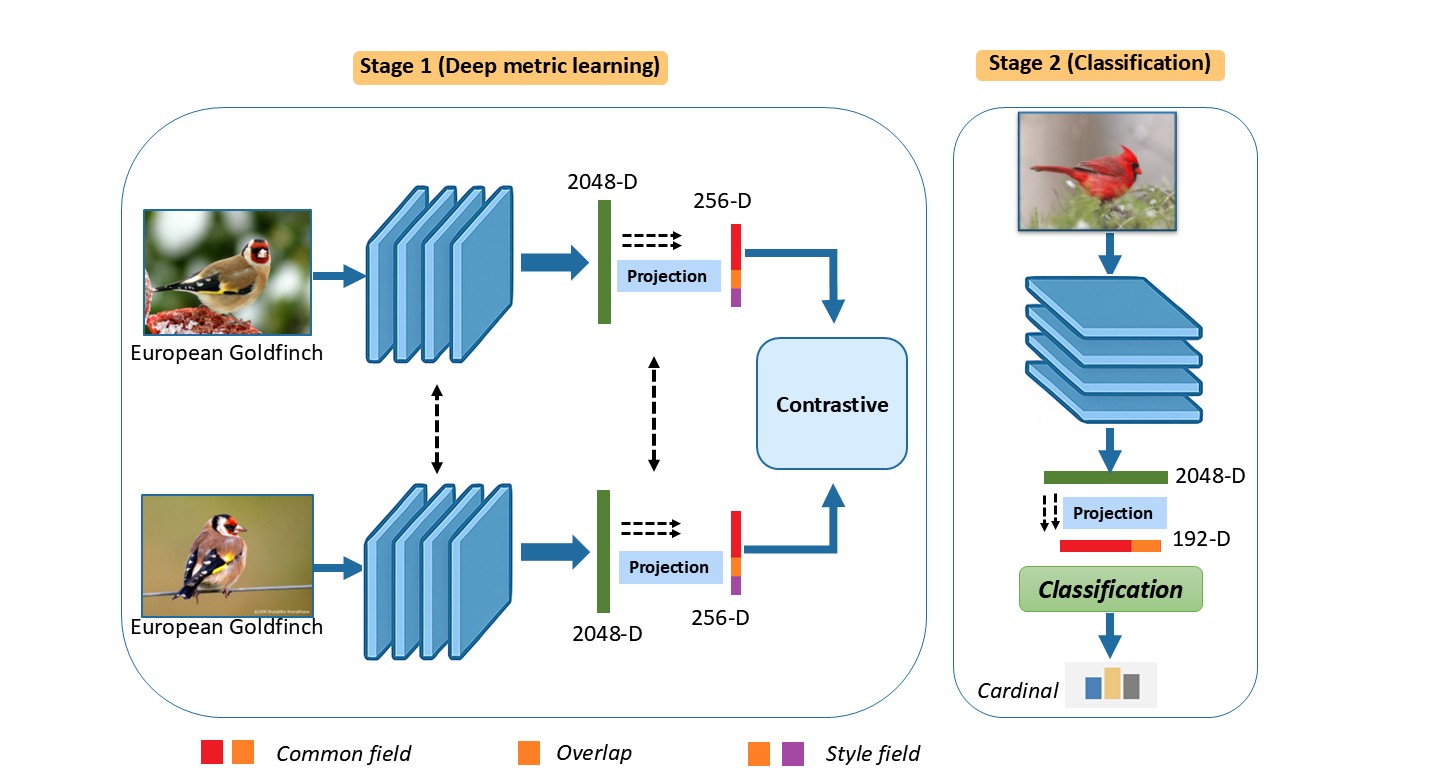}
    \caption{Illustration of CS-SupCon. Stage 1: Deep metric learning explicitly decomposes embeddings into distinct common and style subspaces. Stage 2: A linear classifier is trained solely on common features, thus disregarding style-related variations.}
    \label{fig:CS-SCL}
\end{figure}

Figure~\ref{fig:CS-SCL} illustrates the complete two-stage training strategy of the CS-SupCon framework, explicitly highlighting the decomposition into common and style feature subspaces.

Moreover, the original CS-SupCon method incorporates an optional \textit{overlapping mechanism}, enabling certain dimensions within the deep embeddings to belong concurrently to both common and style subspaces. Such an overlapping strategy grants flexibility in accommodating diverse data distributions and has been empirically demonstrated to improve the generalization capabilities of the model~\cite{dornaika2025deep}. Nevertheless, our proposed SCS-SupCon framework in this study is constructed based on the non-overlapping variant of CS-SupCon, explicitly adopting a sigmoid-based contrastive loss to improve the discrimination between negative pairs.

\subsection{CS-SupCon Loss and Training Procedure}

Let $I$ represent the set of indices for augmented samples within a training batch. For each index $i \in I$, $P(i)$ indicates the set of positive samples that share the same class label as sample $i$. The CS-SupCon framework jointly optimizes the encoder and projection head by minimizing the following composite loss function:
\begin{equation} 
\label{eq:cs_supcon_loss}
\begin{split}
\mathcal{L}_{\text{CS-SupCon}} = 
\sum_{i \in I} \frac{1}{|P(i)|} \sum_{p \in P(i)} 
\bigg\{ 
& - \log \frac{\exp(\cvect_i \cdot \cvect_p / \tau)}{\sum_{j \in I\setminus \{i\}} \exp(\cvect_i \cdot \cvect_j / \tau)} \\
& + \alpha \log \frac{\exp(\svect_i \cdot \svect_p / \tau)}{\sum_{j \in I\setminus \{i\}} \exp(\svect_i \cdot \svect_j / \tau)} 
- \beta \| \svect_i - \svect_p \| 
\bigg\}.
\end{split}
\end{equation}  

Here, the "$\cdot$" symbol denotes the dot product operation. The temperature parameter $\tau$ regulates the scale of similarity scores, while hyperparameters $\alpha$ and $\beta$ control the influence of style-related terms. Specifically, this composite loss consists of three distinct components:

\begin{itemize}
    \item The first term, analogous to the traditional SupCon loss, encourages common feature representations of positive pairs to be compact.
    \item The second term explicitly promotes diversity among the style representations of positive samples, thus increasing style variability.
    \item The third term explicitly imposes a Euclidean distance constraint, enlarging the distances between style features of positive pairs, which reinforces the disentanglement between style and content.
\end{itemize}

\subsection{Classifier Training on Common Features}

After training the encoder and projection head using the CS-SupCon loss defined in Eq.~\eqref{eq:cs_supcon_loss}, their parameters are fixed. Subsequently, in the second stage, a linear classifier is trained solely on the common feature representations $\cvect$ by minimizing the cross-entropy loss:
\begin{equation}
\mathcal{L}_{\text{CE}}(\hat{\yvect}, \yvect) = -\sum_{k=1}^{C} y_k \log(\hat{y}_k),
\end{equation}
where $\hat{\yvect}\in \mathbb{R}^{C}$ denotes the predicted probability distribution over classes, and $\yvect\in \mathbb{R}^{C}$ is the ground truth represented as a one-hot vector. Exclusively training on common features ensures that the classifier concentrates solely on discriminative, class-relevant information, thus improving generalization performance by effectively reducing interference from intra-class style variations.

\section{Proposed Approach: Sigmoid-based Common and Style Supervised Contrastive Learning with Adaptive Decision Boundaries (SCS-SupCon)}

\label{sec4}

In this section, we formally present our proposed framework termed Sigmoid-based Common and Style Supervised Contrastive Learning with Adaptive Decision Boundaries (SCS-SupCon). Our method explicitly integrates a sigmoid-based contrastive loss function featuring adaptive decision boundaries, aiming to achieve robust and effective feature disentanglement. Furthermore, we maintain the explicit style-distance constraint utilized in prior approaches to further strengthen disentanglement performance.

\subsection{Motivation and Overview}

While CS-SupCon explicitly decomposes embeddings into distinct common and style subspaces, its dependence on the conventional InfoNCE-based contrastive loss leads to insufficient discrimination among negative pairs, thereby restricting the robustness of the learned feature representations. To overcome this shortcoming, we introduce the SCS-SupCon framework, which incorporates a novel sigmoid-based contrastive loss with two learnable parameters—temperature ($t=\exp(t')$) and bias ($b$). This approach notably strengthens the differentiation between positive and negative pairs, significantly enhancing the discriminative power and generalization capability of the resulting representations.

\begin{figure}[!htbp]
    \centering
    \includegraphics[width=\textwidth]{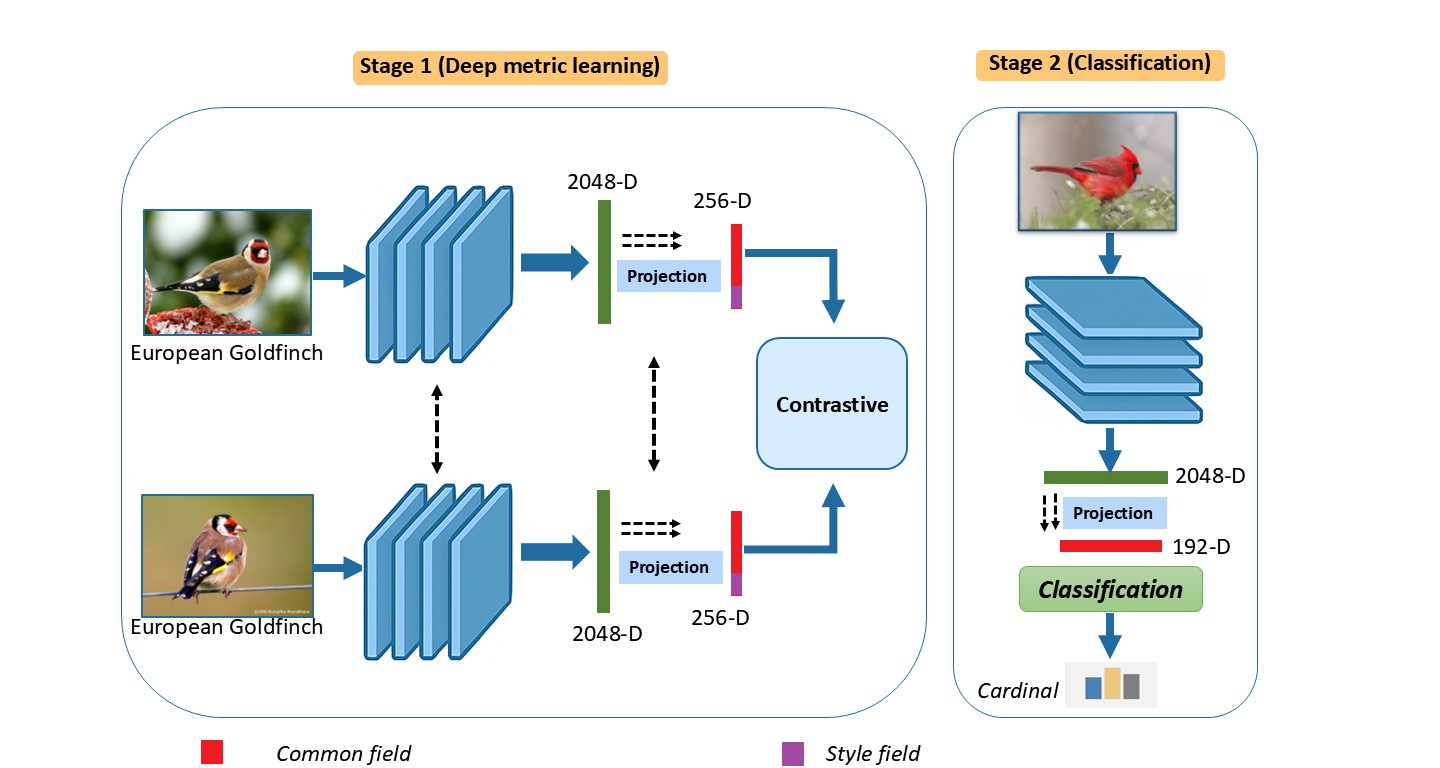}
    \caption{Overview of SCS-SupCon: Sigmoid-based Common and Style Supervised Contrastive Learning with Adaptive Decision Boundaries. Stage 1: Deep metric learning explicitly decomposes embeddings into common and style subspaces, utilizing a sigmoid-based contrastive loss with adaptive decision boundaries. Stage 2: A linear classifier is trained solely on the sigmoid-disentangled common features.}
    \label{fig:SCS-SCL}
\end{figure}

The complete two-stage training strategy of the proposed SCS-SupCon framework is depicted in Figure~\ref{fig:SCS-SCL}. This illustration clearly emphasizes the key distinction from the CS-SupCon approach (Figure~\ref{fig:CS-SCL}), specifically highlighting our integration of the sigmoid-based contrastive loss.

\subsection{Proposed SCS-SupCon Loss Function}

Formally, the proposed SCS-SupCon loss is defined as:
\begin{equation}
\label{eq:scs_supcon_loss_final}
\mathcal{L}_{\text{SCS-SupCon}} = 
-\frac{1}{|I|^2}\sum_{u \in I}\sum_{v \in I}\log\frac{1}{1 + e^{z_{uv}(-\exp(t') \,  \cvect_u \cdot \cvect_v + b)}} 
- \frac{\beta}{|I|}\sum_{i \in I}\frac{1}{|P(i)|}\sum_{p \in P(i)}\|\svect_i-\svect_p\|.
\end{equation}

In the above equation, the indices $u$ and $v$ represent an arbitrary sample pair within the current training batch, explicitly encompassing both positive and negative pairs. \textcolor{black}{The indicator $z_{uv} \in \{+1,-1\}$ encodes the supervised pair label: $z_{uv}=+1$ if samples $u$ and $v$ share the same class and $z_{uv}=-1$ otherwise.} Distinct from traditional supervised contrastive losses, which predominantly focus on anchor–positive pairs, our sigmoid-based contrastive loss explicitly considers the relationships between every labelled pair of samples in the batch. \textcolor{black}{While the distinction between positive and negative pairs is still governed by the supervised labels through $z_{uv}$, the loss evaluates all such pairs inside the mini-batch, leading to a denser and more informative set of supervised pairwise constraints. On the other hand, the InfoNCE loss concentrates on contrasting a given positive pair within a set of samples.} This holistic modeling substantially improves the model’s capability to discriminate subtle distinctions between positive and negative pairs, thus enhancing robustness and generalization performance. In particular, the learnable parameter $t'$ ensures that the temperature parameter $t=\exp(t')$ remains positive and numerically stable.

\begin{figure}[htbp]
    \centering
    \includegraphics[width=1.0\linewidth]{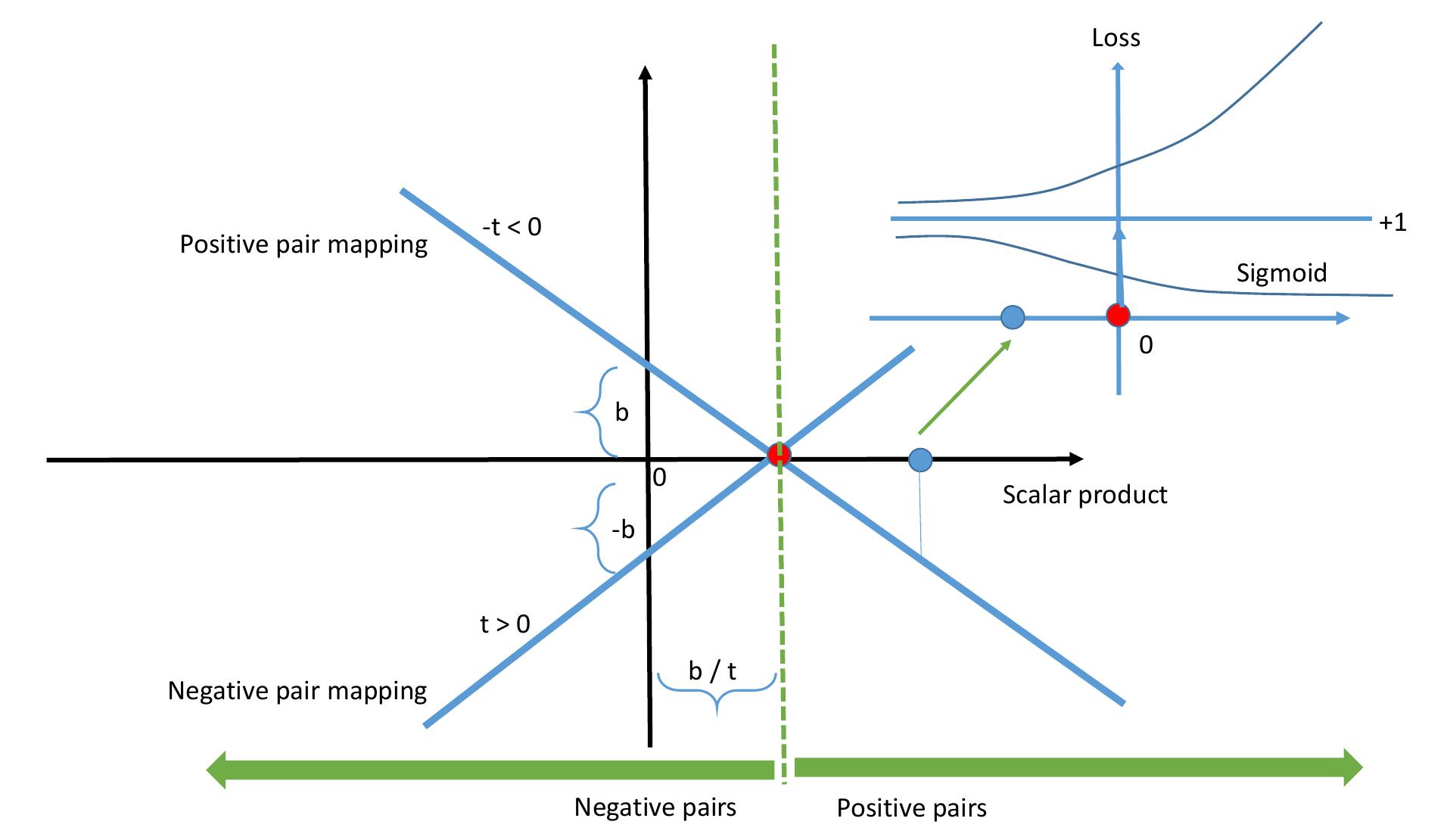}
    \caption{\textcolor{black}{Illustration of the mapping from the
    scalar product $r_{uv} = \cvect_u \cdot \cvect_v$ to the logits of positive and
    negative pairs, the sigmoid output, and the resulting logistic loss. The sigmoid loss encourages learning deep features with the following property: the scalar product of a positive pair should lie beyond the decision boundary (indicated by the red dot), while the scalar product of a negative pair should lie on the opposite side of that boundary. The adaptive boundary itself is  learnable. }}
    \label{fig:sigmoid_mapping}
\end{figure}

\textcolor{black}{Figure~\ref{fig:sigmoid_mapping} provides a complementary geometric view of Eq.~(4). Let $r_{uv} = \cvect_u \cdot \cvect_v$ denote the scalar product between the common features of a pair $(u,v)$. For a positive pair ($z_{uv}=+1$), the argument fed to the sigmoid is an affine function $x_{+}(r_{uv}) = -t\,r_{uv} + b$, whereas for a negative pair ($z_{uv}=-1$) it becomes $x_{-}(r_{uv}) = t\,r_{uv} - b$. The two lines $x_{+}(r_{uv})$ and $x_{-}(r_{uv})$ intersect at $r_{uv}^\star = b/t$, which therefore acts as the decision boundary in the similarity space. These arguments are then passed through a decreasing sigmoid $\sigma(x) = 1/(1+e^{x})$, so that $\sigma(x)\to 1$ as $x\to -\infty$ and $\sigma(x)\to 0$ as $x\to +\infty$. The logistic loss $-\log \sigma(x) = \log(1+e^{x})$ takes values in $[0,+\infty)$: it is close to zero for large negative $x$ and grows without bound as $x$ becomes large and positive. Consequently, high-similarity positive pairs are mapped to negative $x$ and incur very small losses, whereas high-similarity negative pairs are mapped to positive $x$ and incur large losses, receiving strong gradients that push their similarities down. In this picture the learnable temperature $t$ controls how rapidly the two lines cross the boundary at $r_{uv}^\star$, and the bias $b$ shifts this boundary along the similarity axis, yielding an adaptive decision threshold between positive and negative pairs.}

\textcolor{black}{Intuitively, for a positive pair ($z_{uv}=+1$) the logit inside the sigmoid, $-t\,\cvect_u \cdot \cvect_v + b$ where $t = \exp(t')$, is encouraged to become negative so that the pair is assigned a high similarity, whereas for a negative pair ($z_{uv}=-1$) the sign is flipped and the same expression is encouraged to become positive so that the pair is assigned a low similarity. The learnable temperature $t'$ (and hence $t$) controls the slope of this transition around the decision boundary and, consequently, the magnitude of the gradient of the sigmoid-based loss with respect to the similarity. The bias $b$ shifts the boundary along the similarity axis, adapting the point at which a pair changes from being treated as ``similar'' to ``dissimilar''. Compared with the softmax-normalized InfoNCE loss, which induces a fixed global trade-off between positives and negatives in each batch, this parametrization allows the model to learn dataset-dependent decision boundaries and to redistribute gradients towards informative hard positives and hard negatives.}

\textcolor{black}{In all our experiments, $t'$ and $b$ are implemented as two global scalar parameters learned jointly with the encoder and projection head through standard backpropagation. This design keeps the additional parameter overhead negligible while offering a flexible, learnable control over the contrastive decision boundary.}

The learnable bias parameter $b$ introduces adaptive decision boundaries to flexibly differentiate between positive and negative pairs. Specifically, a positive $b$ value biases the model toward classifying pairs as positive, whereas a negative $b$ decreases the threshold for positive pairs identification. Additionally, the style-distance penalty, regulated by the hyperparameter $\beta$, is incorporated to enhance feature disentanglement. \textcolor{black}{It is also worth emphasizing the complementary roles played by the sigmoid-based contrastive term and the style-distance penalty. The first term in Eq.~(\ref{eq:scs_supcon_loss_final}) operates exclusively on the common field $\mathbf{c}$ and is responsible for shaping decision boundaries that maximise inter-class separability while preserving sufficient gradients around hard positive and hard negative pairs. In contrast, the second term acts on the style field $\mathbf{s}$ and explicitly encourages intra-class diversity along style features, thereby discouraging the encoder from encoding class-discriminative information in $\mathbf{s}$. As a result, the two components jointly drive the network to concentrate class-relevant information in the common subspace (where adaptive boundaries are learned by the sigmoid loss) and to relegate nuisance factors (e.g., pose, illumination, texture) to the style subspace. This separation leads to more stable and interpretable decision regions, particularly in fine-grained settings where subtle appearance variations are common.}

\subsection{Two-Stage Training Procedure}

The proposed SCS-SupCon approach employs a clearly defined two-stage training scheme:

\paragraph{Stage 1: Sigmoid-based Feature Disentanglement (Metric Learning)}

In the initial stage, the encoder and projection head are jointly trained by minimizing the proposed SCS-SupCon loss defined in Eq.~\eqref{eq:scs_supcon_loss_final}. The sigmoid-based contrastive loss explicitly enhances the distinction among negative pairs, thereby effectively disentangling the common and style subspaces.

\paragraph{Stage 2: Classifier Training with Sigmoid-Disentangled Common Features}

Upon completing the first training stage, the parameters of both the encoder and projection head are frozen. In the second stage, we train a linear classifier exclusively utilizing the sigmoid-disentangled common feature representations $\cvect$. The classifier optimization is guided by minimizing the conventional cross-entropy loss:
\begin{equation}
\mathcal{L}_{\text{CE}}(\hat{\yvect}, \yvect) = -\sum_{k=1}^{C} y_k \log(\hat{y}_k),
\end{equation}
where $\hat{\yvect}\in\mathbb{R}^{C}$ denotes the predicted class probability distribution, and $\yvect\in\mathbb{R}^{C}$ represents the ground truth labels encoded in a one-hot format. By explicitly excluding style variations and leveraging only common features refined through our sigmoid-based contrastive loss approach, this training strategy significantly strengthens the classifier’s robustness and enhances its generalization capability.

\textcolor{black}{We deliberately adopt a classifier that operates solely on the common field for two reasons. First, the style field is trained to encode intra-class variations (e.g., pose, illumination, background) that are explicitly encouraged to be non-discriminative by the style-distance term; including these features in the final classifier would counteract this objective and, in our preliminary experiments, led to reduced robustness under distribution shifts. Second, simple fusion strategies that concatenate common and style features in Stage~2 did not yield consistent improvements over using the common field alone on our fine-grained benchmarks. We therefore treat the style representation as an auxiliary variable that regularises the common space during Stage~1, while keeping the final decision function focused on the most stable, class-relevant components.}

\paragraph{{\bf Transformer-based Encoder}}

\begin{figure}[htbp]
    \centering
    \includegraphics[width=0.95\textwidth]{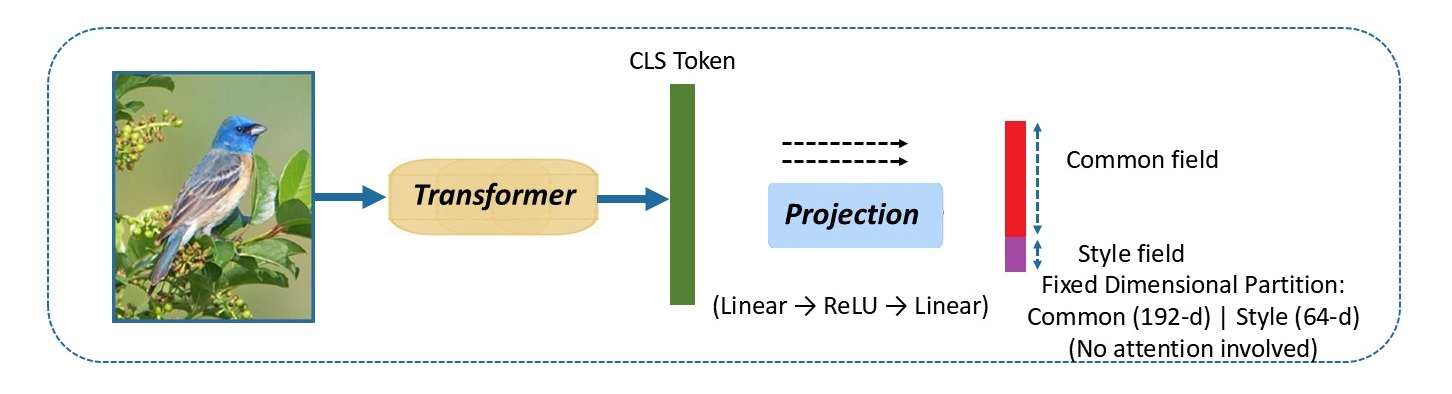}
    \caption{\textcolor{black}{Illustration of deep feature extraction and composition using Transformers within the proposed SCS-SupCon framework. The CLS token produced by the Transformer encoder is first projected through a two-layer MLP with a ReLU activation, and then explicitly partitioned into common (192-d) and style (64-d) subspaces through a fixed dimensional partition without employing attention mechanisms. The sigmoid-based contrastive loss explicitly captures pairwise relationships on the common features, significantly enhancing feature disentanglement, especially beneficial for fine-grained classification tasks.}}
    \label{fig:Transformer_SCS}
\end{figure}

When adopting a Transformer-based encoder in the proposed SCS-SupCon framework, the deep features (CLS token) extracted by the Transformer first undergo projection through a dedicated projection head (implemented as two fully connected layers with a ReLU activation in between). These projected features are then explicitly partitioned into common and style representations, as depicted in Figure~\ref{fig:Transformer_SCS}. Crucially, the SCS-SupCon employs a sigmoid-based contrastive loss to explicitly characterize pairwise relationships among the common features. This strategy notably strengthens the disentanglement capability, thereby significantly improving robustness in fine-grained classification scenarios.
\textcolor{black}{Concretely, the CLS token of dimension $D_{\text{enc}}$ produced by the final transformer block is fed into the two-layer MLP projection head to obtain a $D_p$-dimensional embedding (with $D_p=256$ in our experiments). This embedding is then split deterministically into two contiguous subvectors of dimensions $D_c$ and $D_s$ (with $D_c=192$ and $D_s=64$), which we interpret as the common and style fields, respectively. No additional attention mechanism is introduced at this stage; the decomposition is implemented solely through this learnable MLP projection and fixed index ranges, which keeps the architecture simple and makes the partitioning fully transparent.}

\section{Performance evaluation}
\label{sec5}
\subsection{Datasets, Encoders, and Compared Methods}

We assess the performance of our proposed SCS-SupCon framework using several publicly available image datasets: CIFAR10, CIFAR100, Tiny-ImageNet, CUB200-2011, Stanford Dogs, and PASCAL VOC 2005. Unless explicitly mentioned otherwise, we follow the conventional train/test splits. Detailed descriptions of these datasets are summarized in Table~\ref{tab:datasets}.

\begin{table}[htbp]
\caption{Datasets used in the experiments.} 
\label{tab:datasets}
\centering
  \begingroup
    \color{black}
\resizebox{0.8\textwidth}{!}{
\begin{tabular}{l  m{1.8cm} r  r  r}
\hline
Dataset & Image size & $ \sharp$ classes & \multicolumn{2}{c}{Standard Split} \\
\cline{4-5}
& & & Training Set & Test Set \\
\hline
CIFAR10 & 32 $\times$ 32 & 10 & 50,000 & 10,000 \\
CIFAR100 & 32  $\times$ 32 & 100 & 50,000 & 10,000 \\
Tiny-ImageNet & 64 $\times$ 64 & 200 & 100,000 & 10,000 \\
CUB200-2011 & 224 $\times$ 224 & 200 & 5,994 & 5,794 \\
Stanford Dogs & 224 $\times$  224 & 120 & 12,000 & 8,580 \\
PASCAL VOC 2005 & 224  $\times$ 224 & 4 & 1,843 & 389 \\
\hline
\end{tabular}
}
\endgroup
\end{table}

Several backbone encoders are employed for evaluation. Specifically, we utilize pretrained CNN encoders including ResNet18~\cite{He15_resnet50}, ResNet50~\cite{He15_resnet50}, and ConvNeXt-T~\cite{liu2022convnet}, as well as two pretrained Transformer models from the Model Zoo library, namely TinyViT (5M) and TinyViT (21M). The methods compared in our evaluation include: 
(1) Baseline, comprising the backbone encoder combined with its default classification head; 
(2) SimCLR~\cite{Chen2020}, an unsupervised contrastive learning method; 
(3) BYOL~\cite{Grill2020}, a negative-free self-supervised representation learning method that relies on architectural asymmetry instead of explicit negative pairs;
(4) SupCon~\cite{khosla2020supervised}, a supervised contrastive learning approach; 
(5) SelfCon~\cite{baeSelfContrastiveLearningSingleviewed2022}, a supervised single-view contrastive method leveraging an auxiliary sub-network to minimize reliance on multiview augmentation; 
(6) the recently introduced CS-SupCon~\cite{dornaika2025deep}, explicitly partitioning embeddings into distinct common and style subspaces; 
(7) CS-SupCon w. ov. (CS-SupCon with overlap)~\cite{dornaika2025deep}, extending CS-SupCon by allowing overlap between common and style embedding dimensions;
(8) PCL~\cite{li2021prototypical}, a prototypical contrastive learning approach that introduces prototypes as cluster centers to improve semantic discrimination; 
(9) PaCo~\cite{cui2021parametric}, a hybrid contrastive learning method incorporating learnable parametric class centers; 
(10) CSTCN~\cite{wang2024learning}, a recently introduced method that combines spatio-temporal clustering with contrastive learning;
(11) TimeSCL~\cite{huang2025noise}, a supervised contrastive method with adaptive temporal weighting;
(12) FNCL~\cite{huynh2022boosting}, decoupling feature normalization to reduce negative-sample dilution; 
(13) Circle Loss~\cite{gui2025cross}, introducing angular margin constraints through class prototypes;
(14) DACL~\cite{xu2025few}, employing dynamic adversarial margins for contrastive learning;
(15) CSA-RSIC~\cite{cheng2025csa}, combining contrastive learning with consistency regularization for multimodal tasks; and 
(16) our proposed SCS-SupCon, introducing a novel sigmoid-based contrastive loss equipped with adaptive decision boundaries, which seamlessly integrates an explicit style-distance constraint to effectively enhance feature disentanglement.

\textcolor{black}{In order to keep the empirical comparison focused and interpretable, we restrict our pool of baselines to supervised models whose training objective explicitly contains a contrastive loss term, either within a two-stage pipeline (e.g., SupCon and CS-SupCon) or a single-stage framework (e.g., PaCo, DACL, TimeSCL, FNCL). Purely margin-based classifier losses or adaptive-softmax variants, as well as diffusion-based generative models for fine-grained recognition, are treated as complementary lines of work. They are discussed qualitatively in Sections~\ref{sec2} and~\ref{sec6}, but are not included in our quantitative benchmarks.}

For all encoders and evaluated methods, we adopt a projection head consisting of two fully connected layers with a ReLU activation in between, mapping encoder outputs into a 256-dimensional embedding space ($D_p = 256$). The previously introduced CS-SupCon method~\cite{dornaika2025deep} explicitly divides the embedding dimensions into common ($D_c = 192$) and style ($D_s = 64$) subspaces. Our proposed SCS-SupCon retains identical dimensional configurations to CS-SupCon; however, we replace its traditional contrastive loss with our novel sigmoid-based contrastive loss to enhance feature discrimination capability. Additionally, the style-distance penalty component (controlled by hyperparameter $\beta$) is directly adopted from CS-SupCon~\cite{dornaika2025deep} to explicitly impose constraints on intra-class style variations.

\subsection{Implementation Details}

We employ standard data augmentation techniques, including random scaling, cropping, and horizontal flipping. Model optimization is conducted through stochastic gradient descent (SGD) with momentum set to 0.9, weight decay set to $1\times 10^{-4}$, and a cosine annealing schedule for the learning rate. Batch sizes and learning rates adopted are consistent with those used in previous experiments (as summarized in Table~\ref{tab:hyperparam}). \textcolor{black}{Specifically, we parameterize the temperature as $t = \exp(t')$, where $t'$ is a learnable scalar. We initialize $t'$ to $\log(0.1)$ so that the corresponding initial temperature is $t = 0.1$, and we initialize $b$ to zero. This choice makes the initial scale of pairwise similarities comparable to the temperature used in InfoNCE-based SupCon and CS-SupCon, which stabilizes early training while still allowing the decision boundary to adapt during optimization. Both $t'$ and $b$ are learned jointly with the encoder and projection head by standard backpropagation.} \textcolor{black}{In addition, preliminary experiments varying the learning rates and batch sizes around the values reported in Table~\ref{tab:hyperparam} yielded very similar accuracies, indicating that the method is not overly sensitive to these initial settings in practice.} \textcolor{black}{To further reduce overfitting, we rely on strong data augmentation (random crops, flips, and colour jitter), weight decay in all experiments, and dropout in the classification head for convolutional backbones. The total number of training epochs is fixed to 1000 for Stage~1 and 100 for Stage~2 across all datasets and methods, as summarized in Table~\ref{tab:hyperparam}. For each dataset, we monitor performance on a held-out validation split (and on the validation folds in our CIFAR-100 cross-validation experiment in Section~\ref{sec5}, Table~\ref{tab:cifar100_res50_5fold}) to verify that training has converged and that no significant overfitting occurs within these fixed training schedules.
}\textcolor{black}{Finally, to further analyse the robustness of the initial values of the temperature and bias, we perform an additional Bayesian hyperparameter optimization experiment using Optuna’s TPE sampler; the protocol and results are described in Section~\ref{subsec:bayesopt}.}

\begin{table}[h!]
  \begingroup
    \color{black}
\centering
\caption{Values of the primary hyperparameters adopted by our proposed SCS-SupCon approach. Here, S1 denotes Stage 1, S2 denotes Stage 2, LR refers to the initial learning rate, Ep indicates the number of training epochs, and BS represents the batch size (shown separately for S1 and S2). These hyperparameters are determined empirically based on previous experiments, effectively guaranteeing stable and successful convergence during training across various datasets.}
\label{tab:hyperparam}
\resizebox{0.9\textwidth}{!}{
\begin{tabular}{p{4cm} p{1.5cm} p{1.5cm} p{1.5cm} p{1.5cm} p{1.5cm}}
\hline
Dataset & LR in S1 & Ep. in S1 & LR in S2 & Ep. in S2 & BS \\ \hline
 CIFAR10 & 0.5 & 1000 & 3 & 100 & 1024/512 \\
 CIFAR100 & 0.5 & 1000 & 3 & 100 & 1024/512 \\
 Tiny-ImageNet & 0.5 & 1000 & 3 & 100 & 1024/512\\
 CUB200-2011 & 0.1 & 1000 & 0.1 & 100 & 96/48\\
 Stanford Dogs (TinyViT(5M)) & 0.1 & 1000 & 0.1 & 100 & 96/48\\
 Stanford Dogs (ConvNeXt-tiny) & 0.1 & 1000 & 0.1 & 100 & 96/48\\
 PASCAL VOC 2005 & 0.1 & 1000 & 0.1 & 100 & 96/48\\
\hline
\end{tabular}
}
\endgroup
\end{table}

\subsection{Experimental Results}

The classification accuracy obtained by the proposed SCS-SupCon method, compared to existing contrastive learning approaches across various datasets and backbone encoders, is summarized in Tables~\ref{tab:results} and~\ref{tab:my-table}. It is evident that SCS-SupCon consistently attains state-of-the-art performance, significantly surpassing classical methods (SupCon, SimCLR, BYOL), recent supervised methods (SelfCon, CS-SupCon, TimeSCL, FNCL), clustering-based approaches (PCL, CSTCN, Circle Loss), and hybrid or adversarial contrastive methods (PaCo, DACL, CSA-RSIC). Notably, the incorporation of the sigmoid-based contrastive loss effectively alleviates the negative-sample dilution issue prevalent in conventional contrastive approaches, resulting in remarkable accuracy gains. These improvements are particularly pronounced on fine-grained datasets, such as CUB200-2011 and Stanford Dogs, and are consistently observed across both CNN and Transformer backbones.

\begin{table}[h!]
\centering
\caption{Performance comparison of different methods evaluated on five image datasets using three distinct encoder architectures.}
\label{tab:results}
\resizebox{1.05\textwidth}{!}{
\begin{tabular}{l|cccccc}
\hline
\backslashbox{Method}{Dataset} & CIFAR10 (\%) $\uparrow$ & CIFAR100 (\%) $\uparrow$ & CUB200-2011 (\%) $\uparrow$ & Stanford Dogs (\%) $\uparrow$ & Stanford Dogs (\%) $\uparrow$ & PASCAL VOC (\%) $\uparrow$ \\
\hline
Backbone & ResNet50 & ResNet50 & TinyViT (21M) & TinyViT (5M) & ConvNeXt-tiny & TinyViT (5M) \\ 
\hline
Baseline & 94.9 & 74.8 & 88.7 & 83.4 & 92.1 & 47.8 \\
SimCLR~\cite{Chen2020} & 93.6 & 73.9 & 80.1 & 77.6 & 86.1 & 47.0 \\
BYOL~\cite{Grill2020} & 95.0 & 75.0 & 80.7 & 78.3 & 88.2 & 45.8 \\
SupCon~\cite{khosla2020supervised} & 95.6 & 75.5 & 89.1 & 85.5 & 92.8 & 51.9 \\
SelfCon~\cite{baeSelfContrastiveLearningSingleviewed2022} & 95.6 & 78.1 & - & - & - & - \\
PCL~\cite{li2021prototypical} & 95.7 & 78.5 & 89.3 & 85.8 & 92.9 & 51.1 \\
PaCo~\cite{cui2021parametric} & 95.8 & 78.3 & 89.2 & 85.7 & 92.7 & 51.2 \\ 
CSTCN~\cite{wang2024learning} & 95.8 & 78.7 & 89.4 & 86.0 & 93.0 & 52.6 \\
TimeSCL~\cite{huang2025noise}   & \textbf{95.9} & 78.7 & 89.4 & 85.8 & 92.9 & 51.9 \\
FNCL~\cite{huynh2022boosting}   & 95.7 & 79.1 & 89.3 & 85.6 & 91.9 & 52.3 \\
Circle Loss~\cite{gui2025cross} & 95.6 & 78.5 & 88.7 & 85.4 & 92.1 & 52.1 \\
DACL~\cite{xu2025few}           & 95.8 & 78.7 & 89.5 & 84.8 & 92.4 & 52.2 \\
CSA-RSIC~\cite{cheng2025csa}    & 95.6 & 78.6 & 89.1 & 86.0 & 92.7 & 52.6 \\
CS-SupCon~\cite{dornaika2025deep} & 95.4 & 77.6 & 89.8 & 86.0 & 92.8 & 51.4 \\
CS-SupCon w. ov.~\cite{dornaika2025deep} & 95.7 & 78.4 & 89.8 & 86.2 & 93.1 & 53.3 \\
SCS-SupCon (Ours) & \textbf{95.9} & \textbf{79.2} & \textbf{90.2} & \textbf{86.6} & \textbf{93.8} & \textbf{53.8} \\ 
\hline
\end{tabular}}
\end{table}

\begin{table}[h!]
\centering
\begingroup
\color{black}
\caption{Comparison of classification accuracy for various contrastive learning methods evaluated with ResNet-18 across five datasets.}
\label{tab:my-table}
\resizebox{1.05\textwidth}{!}{
\begin{tabular}{l|ccccc}
\hline
\backslashbox{Method}{Dataset} & CIFAR10 (\%) $\uparrow$ & CIFAR100 (\%) $\uparrow$ & Tiny-ImageNet (\%) $\uparrow$ & CUB200-2011 (\%) $\uparrow$ & Stanford Dogs (\%) $\uparrow$ \\ \hline
Baseline & 94.7 & 72.9 & 57.5 & 55.9 & 61.5 \\
SimCLR~\cite{Chen2020} & 93.0 & 72.3 & 55.9 & 55.2 & 60.8 \\
BYOL~\cite{Grill2020} & 94.6 & 73.0 & 57.2 & 56.2 & 62.3 \\
SupCon~\cite{khosla2020supervised} & 94.7 & 73.6 & 57.7 & 57.1 & 63.0 \\
SelfCon~\cite{baeSelfContrastiveLearningSingleviewed2022} & 95.1 & 74.9 & 59.8 & 60.4 & 65.4 \\
PCL~\cite{li2021prototypical} & 95.1 & 75.0 & 60.0 & 60.5 & 65.6 \\
PaCo~\cite{cui2021parametric} & 95.2 & 74.8 & 59.7 & 60.2 & 65.2 \\
CSTCN~\cite{wang2024learning} & 95.3 & 75.2 & 60.3 & 61.2 & 66.7 \\
TimeSCL~\cite{huang2025noise} & 95.2 & 75.1 & 59.4 & 60.5 & 66.6 \\
FNCL~\cite{huynh2022boosting} & 95.3 & 74.8 & 60.1 & 61.0 & 66.4 \\
Circle Loss~\cite{gui2025cross} & 95.3 & 75.0 & 60.3 & 60.9 & 66.5 \\
DACL~\cite{xu2025few} & \textbf{95.4} & 74.3 & 60.2 & 60.7 & 66.7 \\
CSA-RSIC~\cite{cheng2025csa} & 95.1 & 74.6 & 60.1 & 60.7 & 66.2 \\
CS-SupCon~\cite{dornaika2025deep} & 94.9 & 74.1 & 59.0 & 59.1 & 64.8 \\
CS-SupCon w. ov.~\cite{dornaika2025deep} & 95.2 & 75.2 & 59.9 & 61.0 & 66.3 \\ 
SCS-SupCon (Ours) & \textbf{95.4} & \textbf{75.5} & \textbf{60.7} & \textbf{62.1} & \textbf{67.6} \\ 
\hline
\end{tabular}
}
\endgroup
\end{table}

Further rigorous validation conducted through five-fold cross-validation on CIFAR-100 (Table~\ref{tab:cifar100_res50_5fold}) clearly indicates that the proposed SCS-SupCon consistently provides statistically significant performance gains over the most closely related supervised contrastive approaches, including SupCon, SelfCon, and particularly the recently proposed CS-SupCon family of methods upon which our method directly builds. Specifically, SCS-SupCon achieves approximately 3.9\% higher mean accuracy than SupCon, around 1.3\% higher than SelfCon, and about 1.7\% higher than the original CS-SupCon method. The statistical significance of these improvements was verified using a paired Student's t-test ($p<0.05$), with statistically significant differences across individual folds indicated by boldface numbers marked with a dagger ($^\dagger$). Note that the Mean column is presented solely for summarization and is not subject to statistical testing. These results further demonstrate that our SCS-SupCon framework, which integrates the sigmoid-based pairwise contrastive loss with the style-distance penalty, significantly enhances discriminative capability, robustness, and generalization compared to these closely related baseline methods.

\begin{table}[h]
\caption{Top-1 accuracy (\%) obtained through five-fold cross-validation on CIFAR-100 using ResNet-50. Results achieving statistically significant improvements over other methods (paired t-test, $p < 0.05$) are indicated with a dagger ($^\dagger$).}
\label{tab:cifar100_res50_5fold}
\resizebox{\textwidth}{!}{
\begin{tabular}{l|ccccc|c}
\hline
Method & Fold-1 (\%) $\uparrow$ & Fold-2 (\%) $\uparrow$ & Fold-3 (\%) $\uparrow$ & Fold-4 (\%) $\uparrow$ & Fold-5 (\%) $\uparrow$ & Mean (\%) $\uparrow$ \\ \hline
SupCon~\cite{khosla2020supervised} & 74.8 & 74.6 & 74.5 & 74.7 & 74.4 & 74.6 \\
SelfCon~\cite{baeSelfContrastiveLearningSingleviewed2022} & 76.9 & 77.3 & 77.1 & 77.2 & 77.5 & 77.2 \\
CS-SupCon~\cite{dornaika2025deep} & 76.6 & 76.9 & 76.7 & 76.8 & 77.0 & 76.8 \\
CS-SupCon w. ov.~\cite{dornaika2025deep} & 78.0 & 77.7 & 77.9 & 77.8 & 77.6 & 77.8 \\ 
SCS-SupCon (Ours) & \textbf{78.4}$^\dagger$ & \textbf{78.6}$^\dagger$ & \textbf{78.7}$^\dagger$ & \textbf{78.3}$^\dagger$ & \textbf{78.5}$^\dagger$ & \textbf{78.5} \\
\hline
\multicolumn{7}{l}{$^\dagger$: Statistically significant improvement ($p<0.05$) over SupCon, SelfCon, CS-SupCon, and CS-SupCon w. ov.}
\end{tabular}}
\end{table}

\subsection{\textcolor{black}{Computational Efficiency Analysis}}
\textcolor{black}{
To evaluate the scalability of SCS-SupCon, we compare its computational cost against SupCon and CS-SupCon on CIFAR-100 using a ResNet-18 encoder, an input resolution of $32\times 32$, and a batch size of 1024, all trained on a single NVIDIA RTX~4090 GPU. 
Table~\ref{tab:efficiency} reports the training time per epoch and the peak GPU memory usage, measured under identical implementation and hardware settings. All three methods share exactly the same encoder and 2-layer MLP projection head; the only difference lies in the contrastive loss.}

\textcolor{black}{The results show that SCS-SupCon incurs only a marginal overhead compared to CS-SupCon and the original SupCon: the epoch time increases slightly and the peak memory footprint remains very close to that of the InfoNCE-based variants. We note that the overlapping variant of CS-SupCon has a very similar computational profile to the non-overlapping version used here, since it only changes the allocation of embedding dimensions; including it in Table~\ref{tab:efficiency} would not affect the conclusion that SCS-SupCon has essentially the same time and memory requirements as CS-SupCon. This behavior is expected, since SCS-SupCon reuses the same pairwise similarity matrix as InfoNCE and introduces only two additional scalar parameters $(t', b)$ in the contrastive loss. In practice, SCS-SupCon can therefore be viewed as a drop-in replacement for InfoNCE-based supervised contrastive losses in terms of computational requirements.}

\textcolor{black}{From a theoretical standpoint, the per-batch computational complexity of SCS-SupCon remains $\mathcal{O}(B^2 D_p)$ for a batch size $B$ and embedding dimension $D_p$, matching the complexity of standard InfoNCE-based supervised contrastive losses. All three losses require computing a full $B \times B$ pairwise similarity matrix among $D_p$-dimensional embeddings and then applying element-wise non-linearities (softmax for InfoNCE or sigmoid for SCS-SupCon). The additional style branch in CS-SupCon and the style-distance term in both CS-SupCon and SCS-SupCon only change constant factors but do not alter the overall $\mathcal{O}(B^2 D_p)$ time and $\mathcal{O}(B^2)$ memory complexity.}

\textcolor{black}{For CS-SupCon and SCS-SupCon, the common and style subspaces have dimensions
$D_c = 192$ and $D_s = 64$, so the style-related terms contribute at most $\mathcal{O}(B^2 D_s)$ and the overall order remains $\mathcal{O}(B^2(D_c + D_s)) = \mathcal{O}(B^2 D_p)$.}

\begin{table}[htbp]
\centering
\caption{\textcolor{black}{Computational efficiency comparison on CIFAR-100 with a ResNet-18 encoder (input $32\times 32$, batch size 1024). Training time is measured per epoch and peak GPU memory is measured in GB.}}
\label{tab:efficiency}
\resizebox{0.7\textwidth}{!}{
\begin{tabular}{lcc}
\hline
\textcolor{black}{Method} & \textcolor{black}{Time per epoch (s)} & \textcolor{black}{Peak GPU memory (GB)} \\
\hline
\textcolor{black}{SupCon~\cite{khosla2020supervised}}       & \textcolor{black}{17.5} & \textcolor{black}{7.1} \\
\textcolor{black}{CS-SupCon~\cite{dornaika2025deep}}        & \textcolor{black}{17.4} & \textcolor{black}{7.2} \\
\textcolor{black}{SCS-SupCon}                               & \textcolor{black}{17.7} & \textcolor{black}{7.1} \\
\hline
\end{tabular}}
\end{table}

\subsection{\textcolor{black}{Visualization of Learned Feature Embeddings}}
\begin{figure}[htbp]
  \centering
  \begin{subfigure}{0.44\textwidth}
    \includegraphics[width=\linewidth]{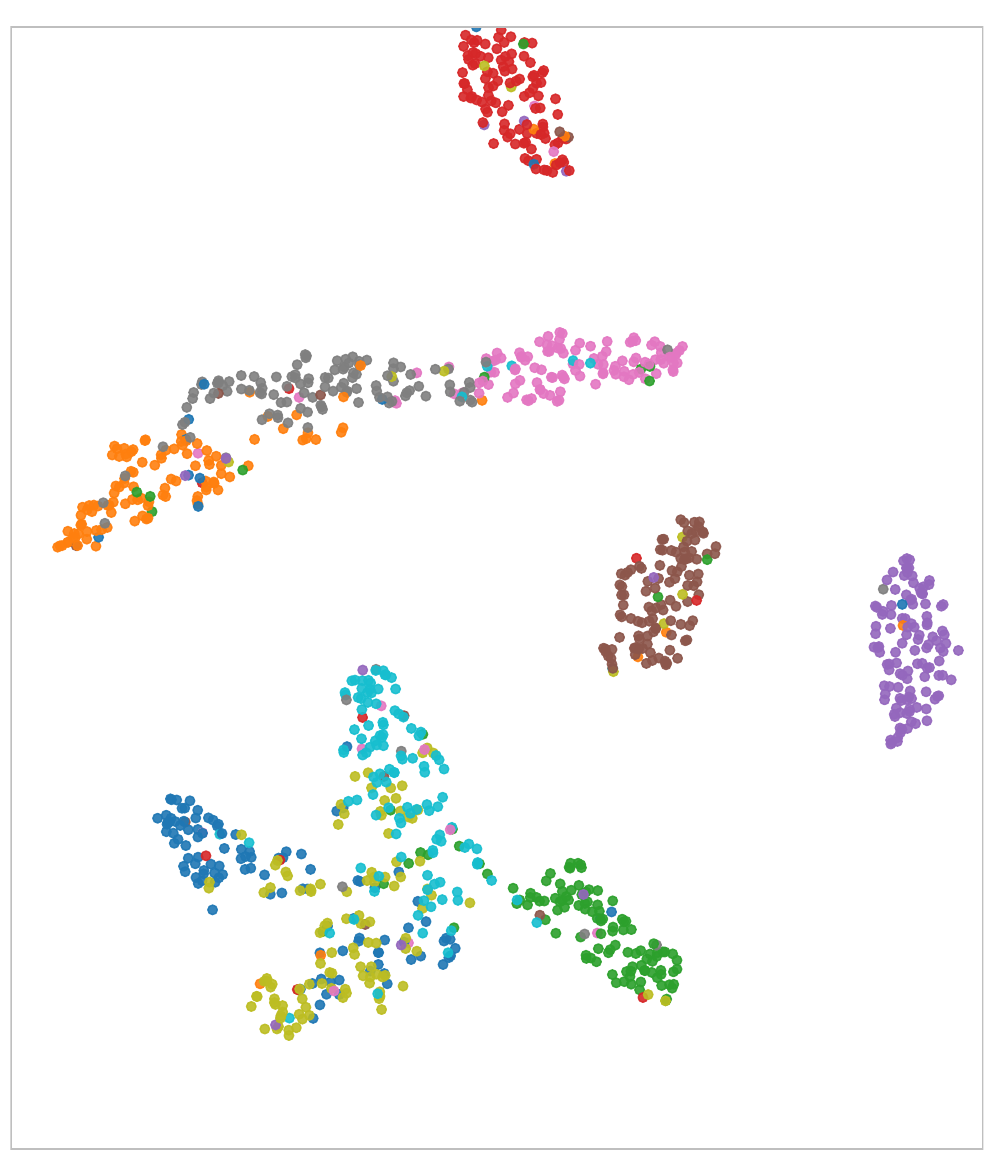}
    \caption{CS-SupCon}
  \end{subfigure}
  \begin{subfigure}{0.44\textwidth}
    \includegraphics[width=\linewidth]{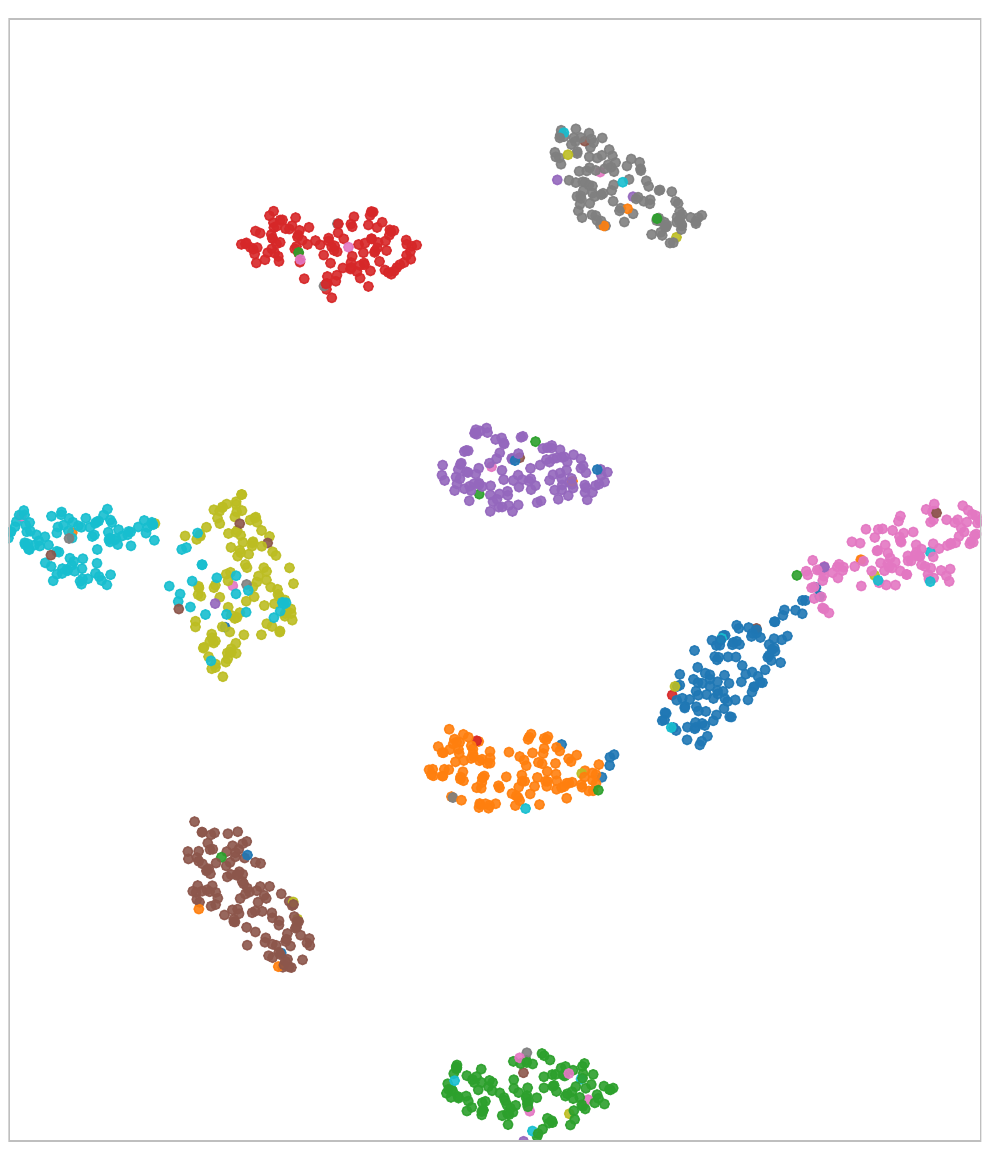}
    \caption{SCS-SupCon}
  \end{subfigure}
  \caption{\textcolor{black}{t-SNE visualization of common-feature embeddings on CUB200-2011 with TinyViT-21M. 
For clarity, we visualize 10 randomly selected bird categories.}}

  \label{fig:tsne}
\end{figure}

\textcolor{black}{To further illustrate the effect of the proposed sigmoid-based contrastive loss and adaptive decision boundaries, we visualize the learned feature embeddings using t-SNE. Figure~\ref{fig:tsne} compares the common-feature embeddings produced by CS-SupCon and SCS-SupCon on the CUB200-2011 dataset with a TinyViT-21M encoder. Compared with CS-SupCon, SCS-SupCon yields tighter intra-class clusters and clearer separation between visually similar fine-grained categories. Several confusing bird species that strongly overlap under CS-SupCon become substantially better separated under SCS-SupCon. These visualizations are consistent with our design goal of alleviating negative-sample dilution and learning more adaptive decision boundaries.}

\subsection{Learnable Parameters and Sensitivity Analysis}
\label{sec:learnable_params}

\begin{figure}[htbp]
    \centering
    \begin{subfigure}[t]{0.48\textwidth}
        \centering
        \includegraphics[width=\linewidth]{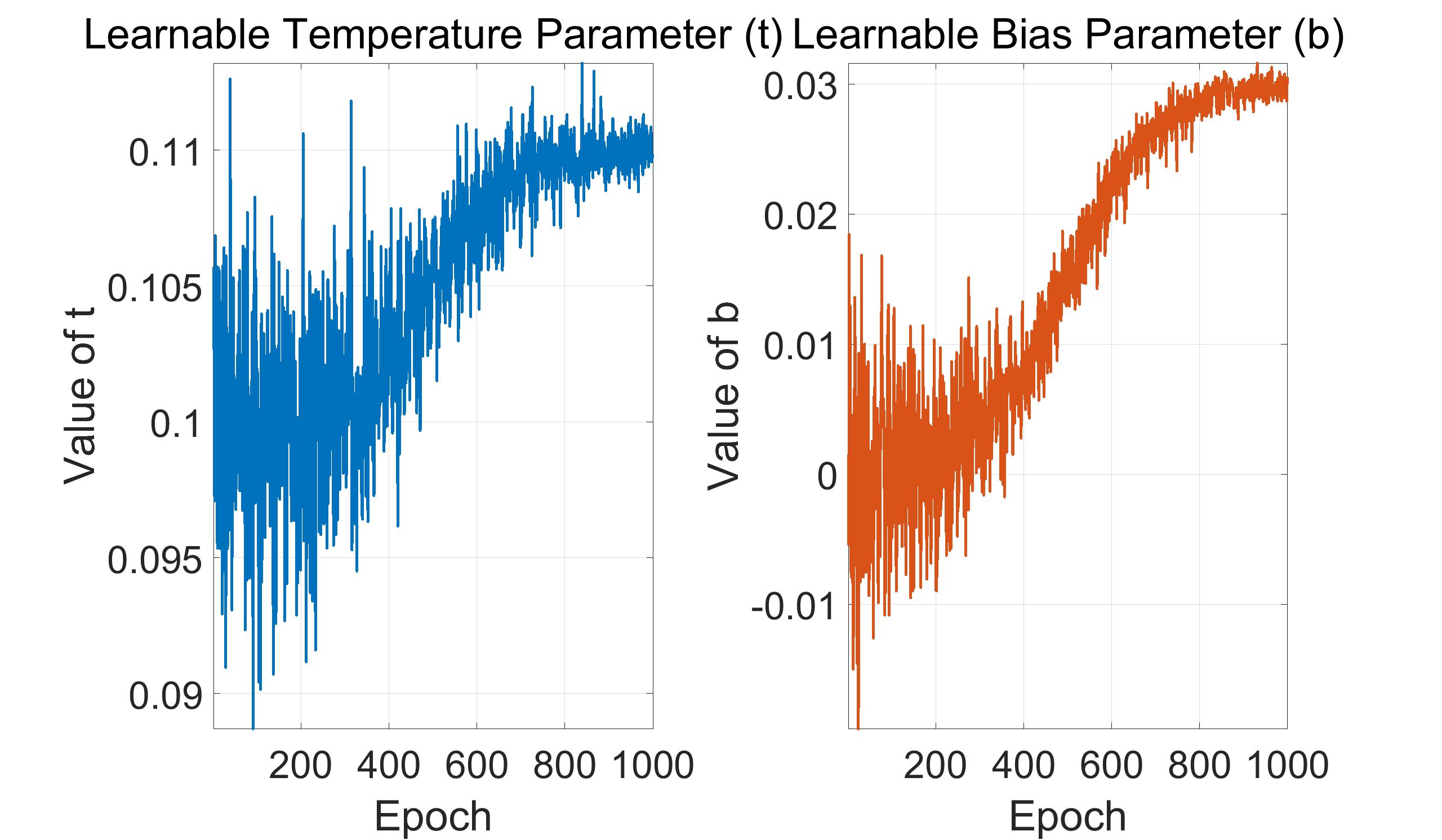}
        \caption{CIFAR-10 (ResNet-50)}
        \label{fig:cifar10_curves}
    \end{subfigure}
    \hfill
    \begin{subfigure}[t]{0.48\textwidth}
        \centering
        \includegraphics[width=\linewidth]{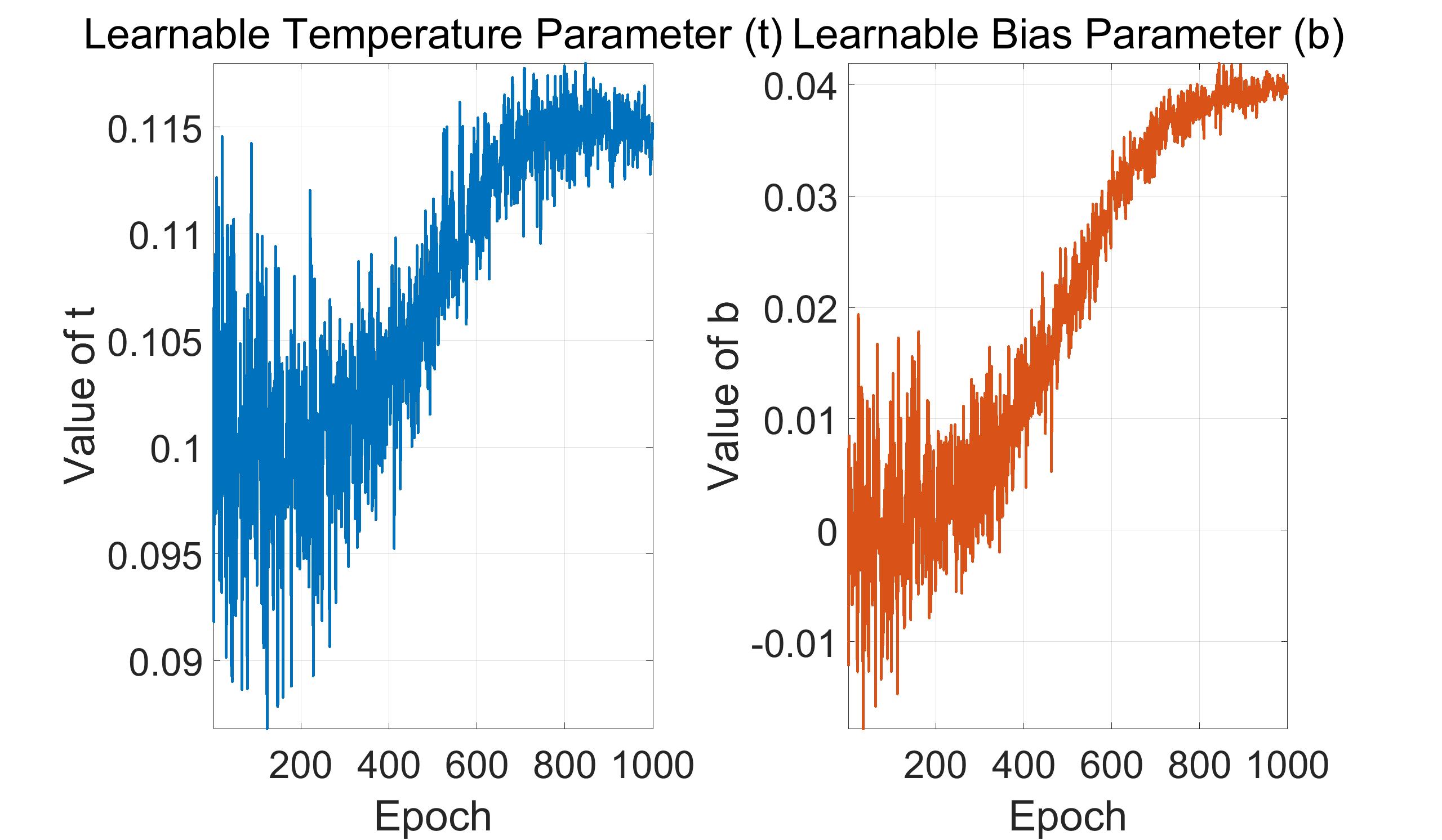}
        \caption{CIFAR-100 (ResNet-50)}
        \label{fig:cifar100_curves}
    \end{subfigure}
    
    \vspace{0.5em}
    
    \begin{subfigure}[t]{0.48\textwidth}
        \centering
        \includegraphics[width=\linewidth]{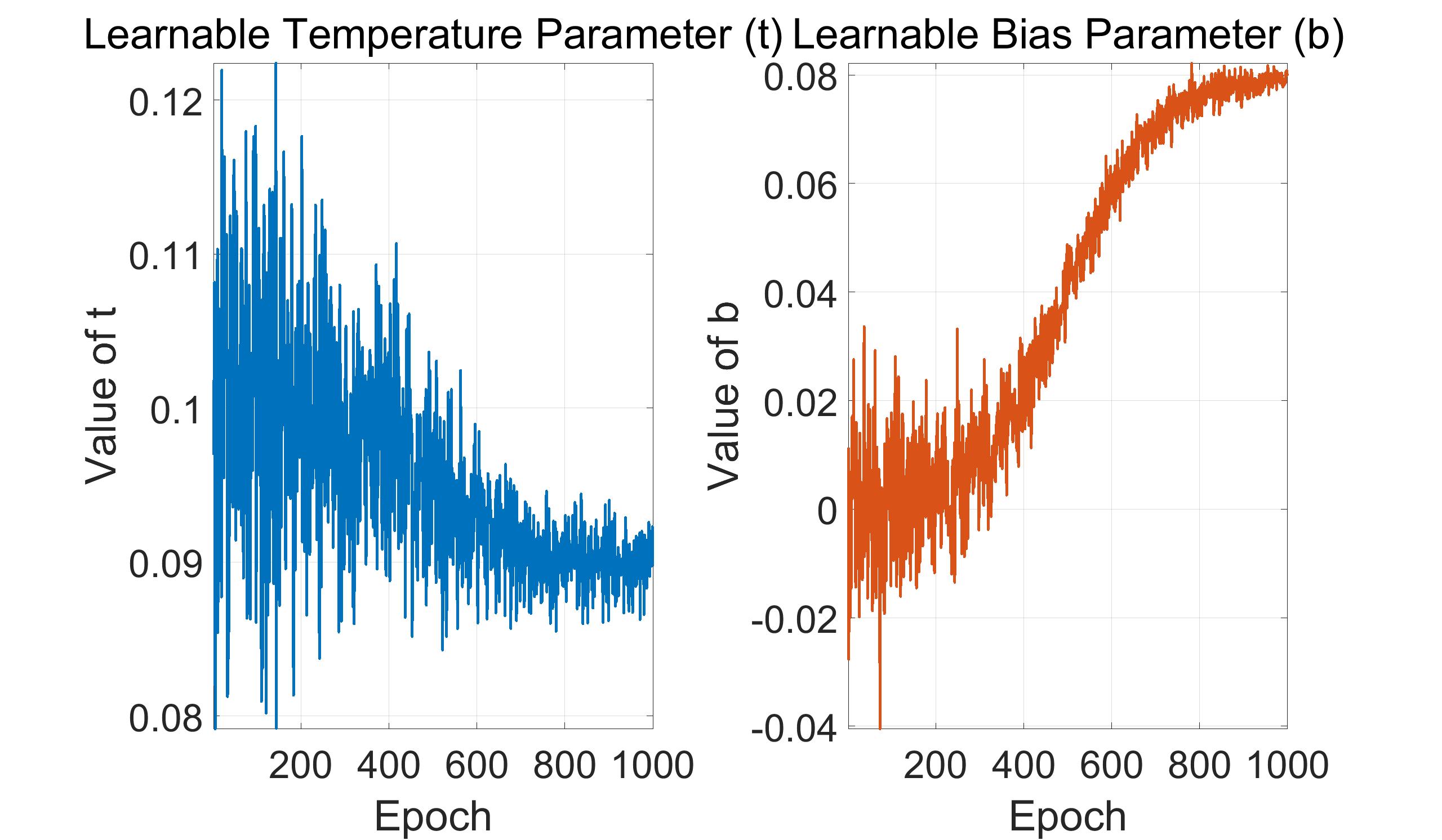}
        \caption{CUB200-2011 (TinyViT-21M)}
        \label{fig:cub_curves}
    \end{subfigure}
    \hfill
    \begin{subfigure}[t]{0.48\textwidth}
        \centering
        \includegraphics[width=\linewidth]{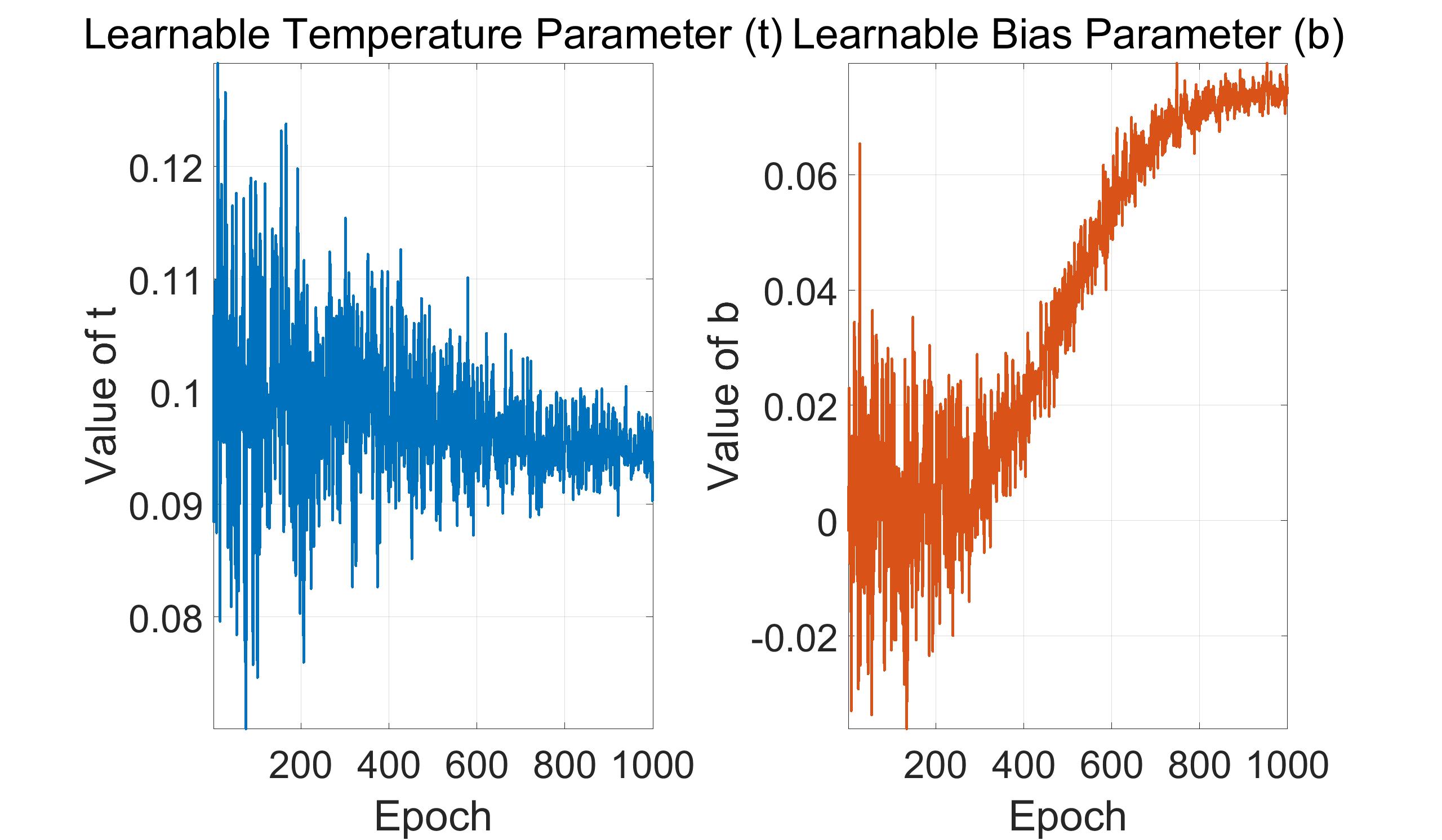}
        \caption{Stanford Dogs (TinyViT-5M)}
        \label{fig:dogs_tinyvit_curves}
    \end{subfigure}
    
    \vspace{0.5em}
    
    \begin{subfigure}[t]{0.48\textwidth}
        \centering
        \includegraphics[width=\linewidth]{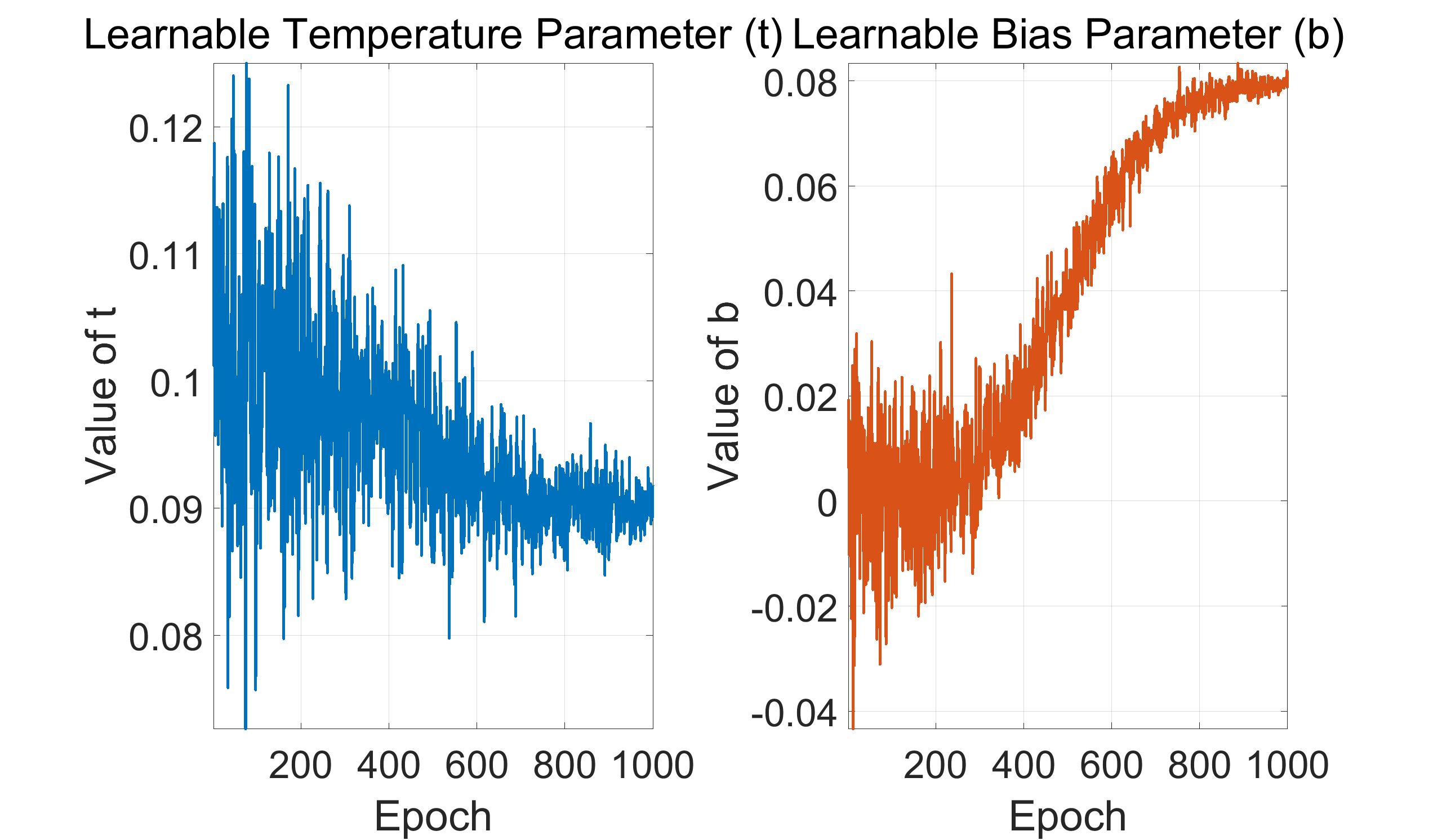}
        \caption{Stanford Dogs (ConvNeXt-tiny)}
        \label{fig:dogs_convnext_curves}
    \end{subfigure}
    \hfill
    \begin{subfigure}[t]{0.48\textwidth}
        \centering
        \includegraphics[width=\linewidth]{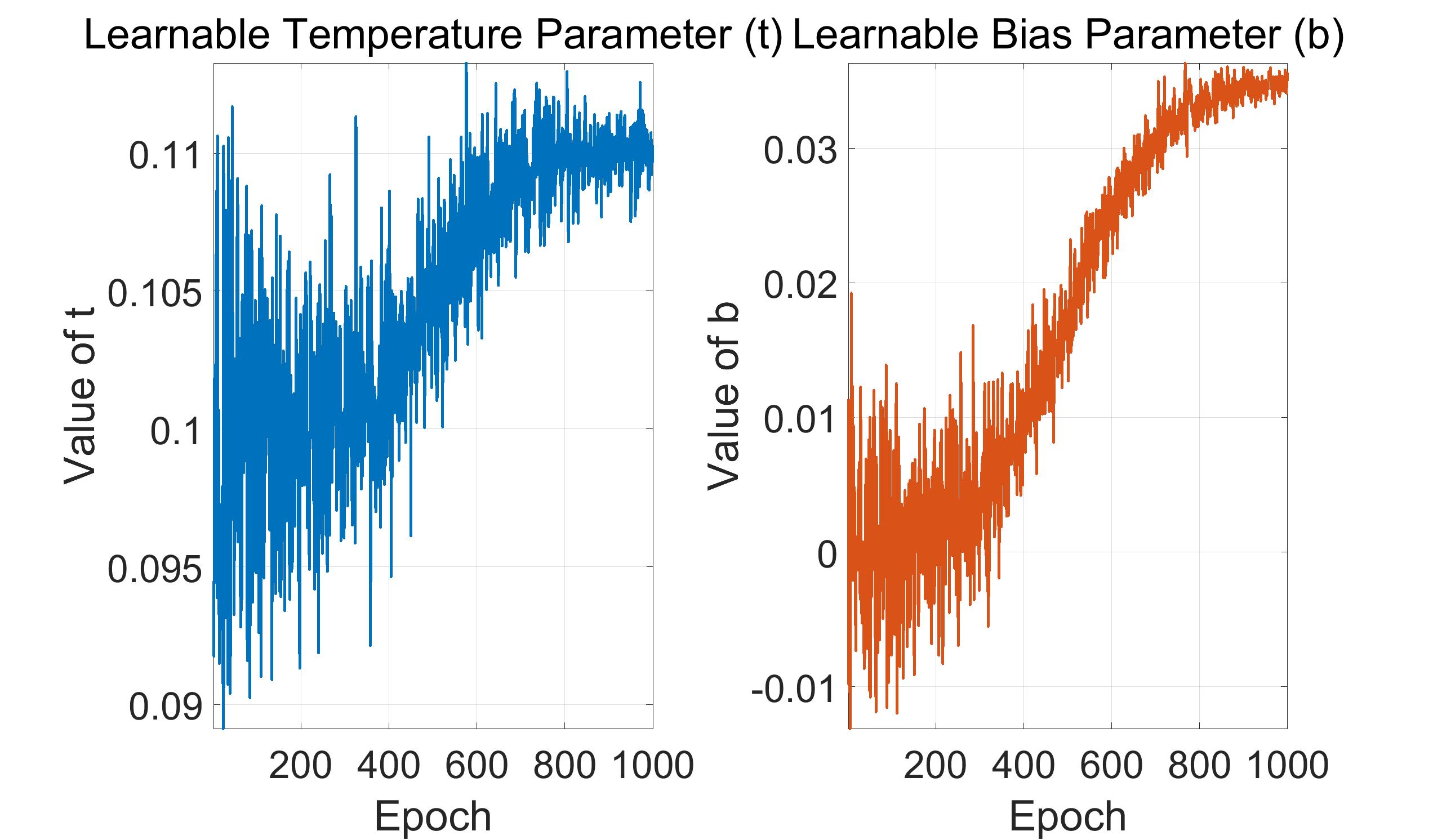}
        \caption{PASCAL VOC 2005 (TinyViT-5M)}
        \label{fig:voc_curves}
    \end{subfigure}
    
    \caption{Across multiple datasets and encoder architectures, the learning curves exhibit uniform convergence trends. These results indicate that the proposed S-SupCon method is both robust and adaptable to diverse learning scenarios. }
    \label{fig:all_curves}
\end{figure}

We analysed the learning behavior of the learnable parameters (temperature $t$, bias $b$) and performed a detailed sensitivity analysis for the hyperparameter $\beta$ employed by the proposed SCS-SupCon method. The training dynamics of these parameters across multiple datasets and backbone architectures are illustrated in Figure~\ref{fig:all_curves}.

\paragraph{Temperature Parameter $t$:}
As depicted in Figure~\ref{fig:all_curves}, the learnable temperature parameter $t$ adaptively converges to distinct stable values depending on the dataset granularity. Specifically, for general datasets (e.g., CIFAR-10, CIFAR-100, PASCAL VOC), the temperature consistently stabilizes at relatively higher values ($0.11$--$0.115$), corresponding to smoother decision boundaries favorable for differentiating broadly separated classes. In contrast, fine-grained datasets (e.g., CUB200-2011, Stanford Dogs) exhibit convergence at lower temperature values ($0.09$--$0.095$), reflecting sharper decision boundaries required to discriminate subtle intra-class differences. Such adaptive convergence behavior further confirms the effectiveness and flexibility of our learnable temperature parameter.

\paragraph{Bias Parameter $b$:}
Likewise, the bias parameter $b$ demonstrates clear adaptive convergence behaviors across various datasets and encoders, as depicted in Figure~\ref{fig:all_curves}. At early training stages, parameter $b$ typically exhibits noticeable fluctuations around zero, reflecting the model's initial uncertainty in setting appropriate decision boundaries. As training progresses, fine-grained datasets (e.g., Stanford Dogs, CUB200-2011) exhibit a consistent and pronounced convergence towards higher positive values (approximately $0.08$), aligning perfectly with our theoretical expectation that positive $b$ values lower the classification threshold for positive pairs. This effectively enhances the model's sensitivity toward subtle intra-class distinctions commonly encountered in fine-grained classification scenarios. Conversely, general-purpose datasets (e.g., CIFAR-10, CIFAR-100, PASCAL VOC) stabilize around lower bias values closer to zero, indicating that minimal adjustments to decision boundaries are sufficient due to inherently clearer inter-class separations. Such diverse yet consistent convergence behaviors across datasets highlight the robustness and flexibility of our proposed adaptive decision boundary mechanism.

\begin{figure*}[htbp]
  \centering
  \subfloat[CIFAR-10 — ResNet-50\label{fig:cifar10_beta}]{
      \includegraphics[width=0.45\textwidth]{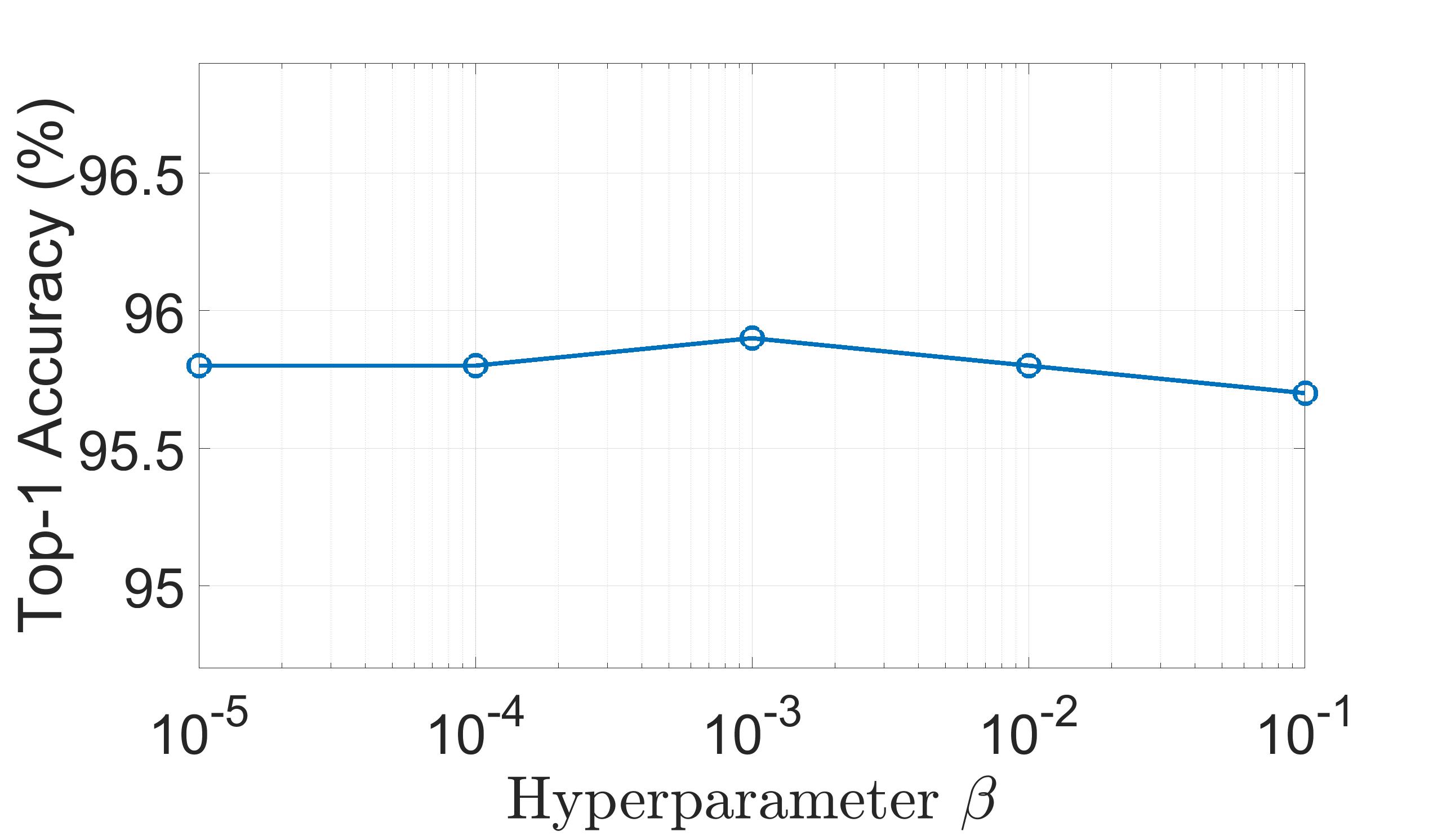}}
  \hfill
  \subfloat[CIFAR-100 — ResNet-50\label{fig:cifar100_beta}]{
      \includegraphics[width=0.45\textwidth]{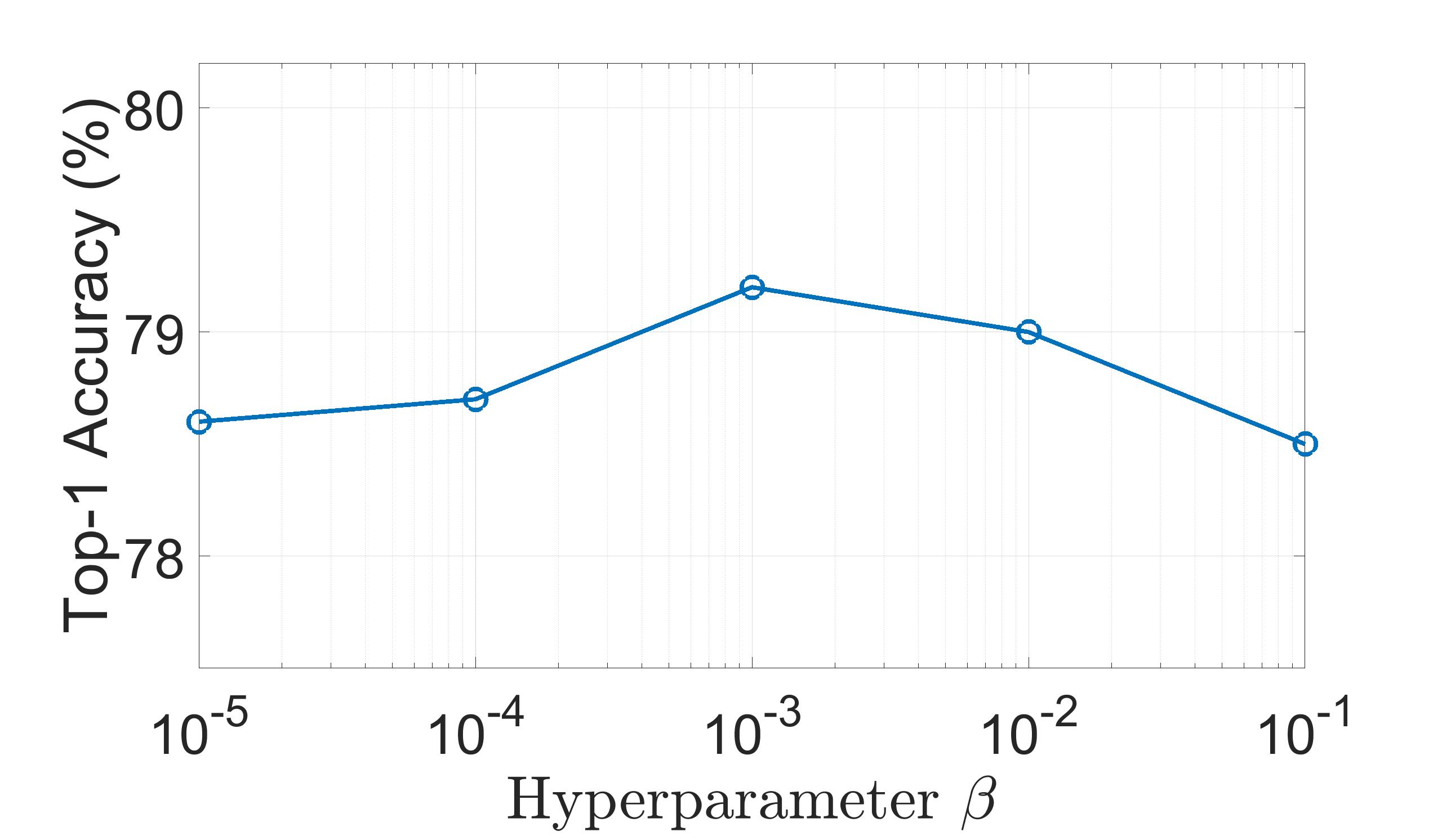}}

  \vspace{0.4em}
  \subfloat[CUB200-2011 — TinyViT-21M\label{fig:cub200_beta}]{
      \includegraphics[width=0.45\textwidth]{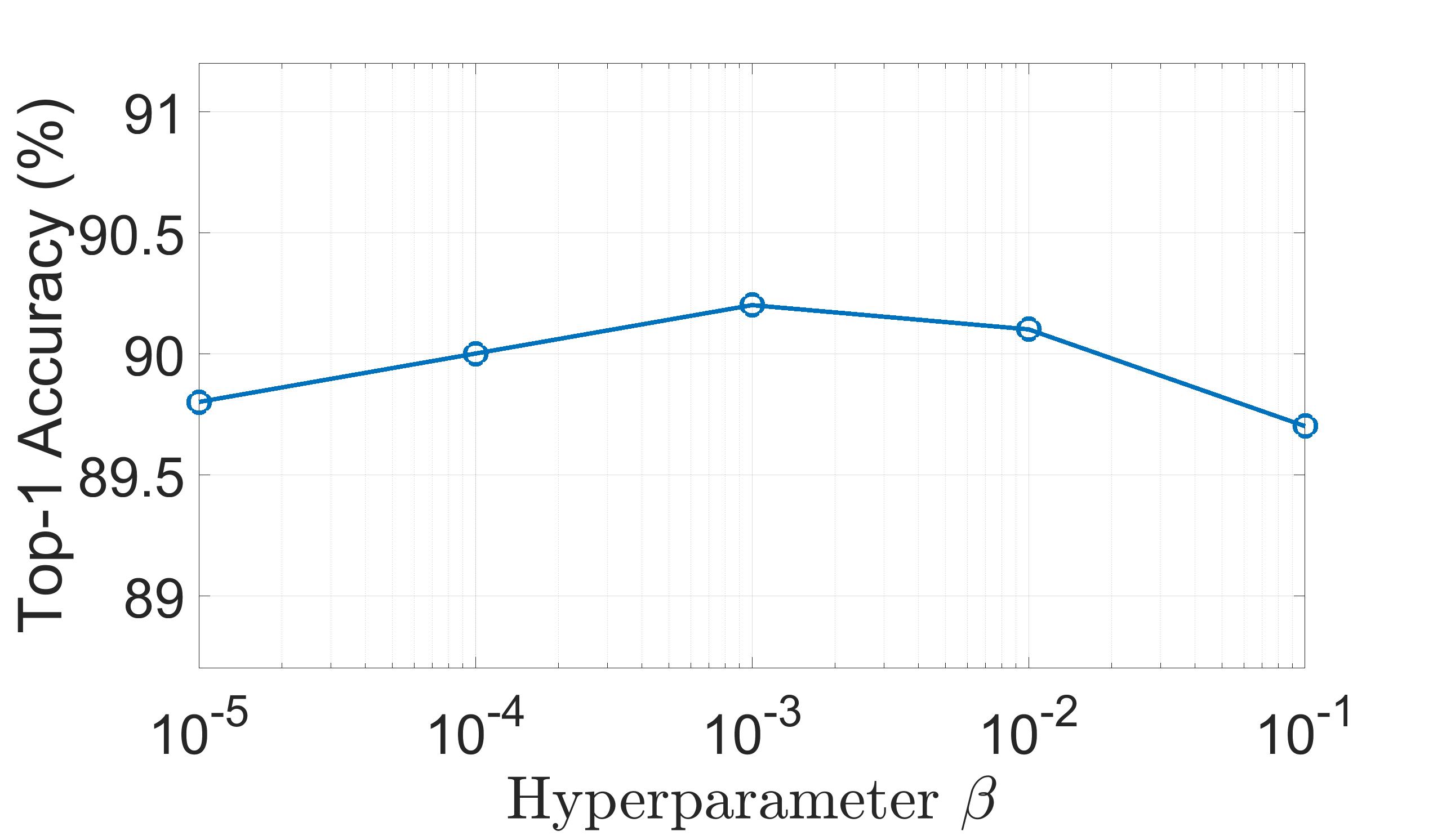}}
  \hfill
  \subfloat[Stanford Dogs — TinyViT-5M\label{fig:dogs_tvit_beta}]{
      \includegraphics[width=0.45\textwidth]{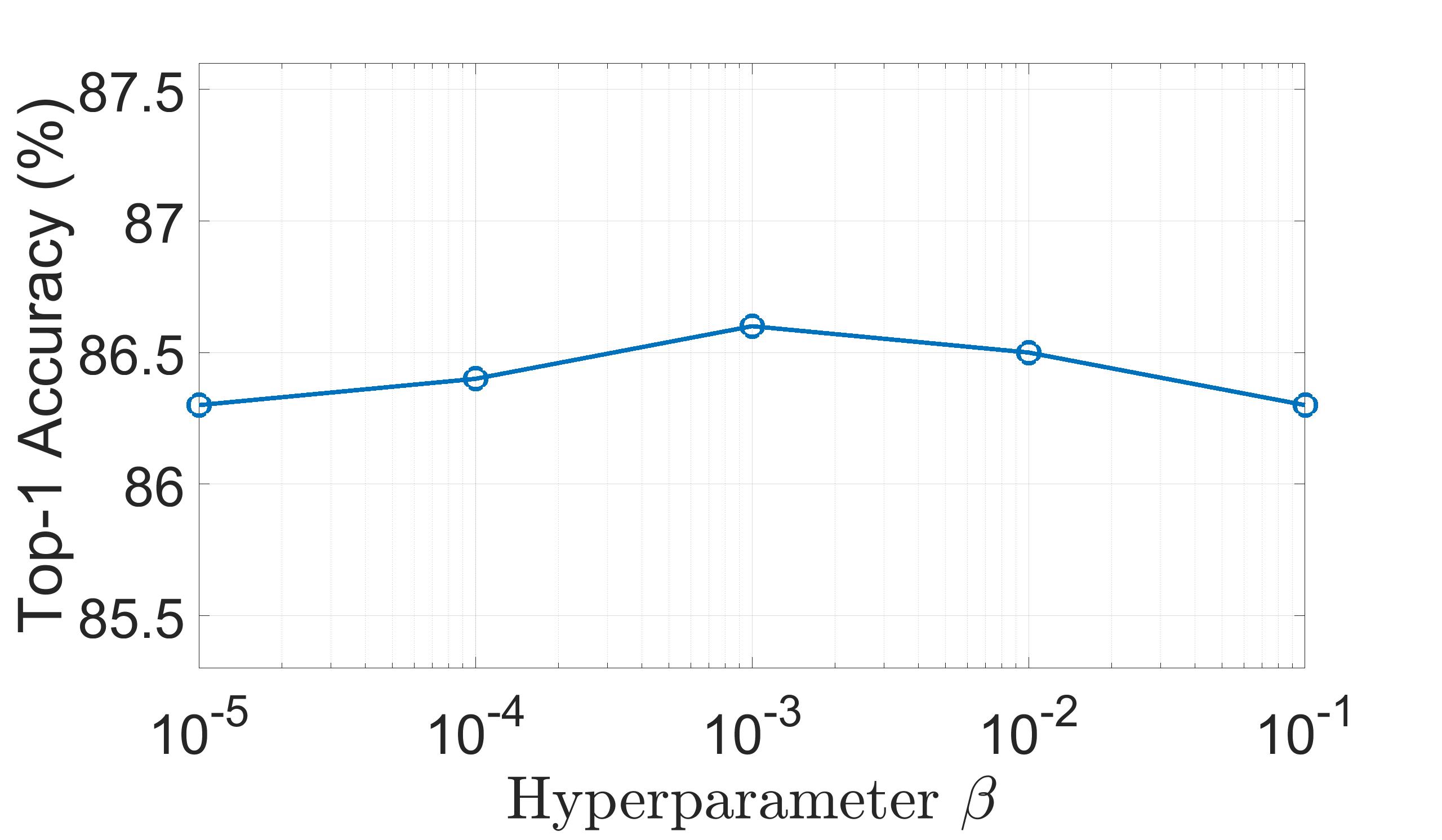}}

  \vspace{0.4em}
  \subfloat[Stanford Dogs — ConvNeXt-Tiny\label{fig:dogs_cnext_beta}]{
      \includegraphics[width=0.45\textwidth]{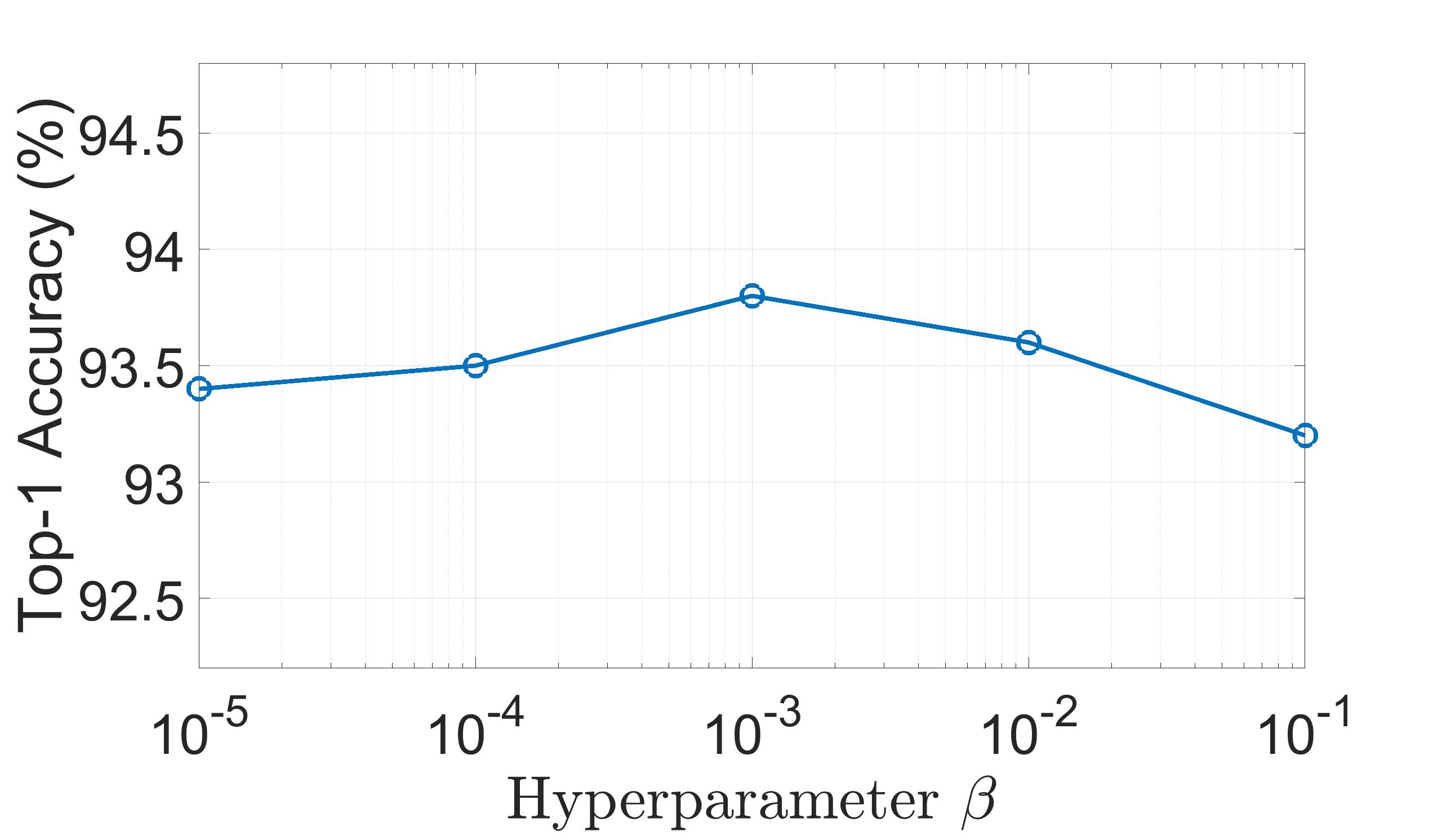}}
  \hfill
  \subfloat[PASCAL VOC — TinyViT-5M\label{fig:voc_beta}]{
      \includegraphics[width=0.45\textwidth]{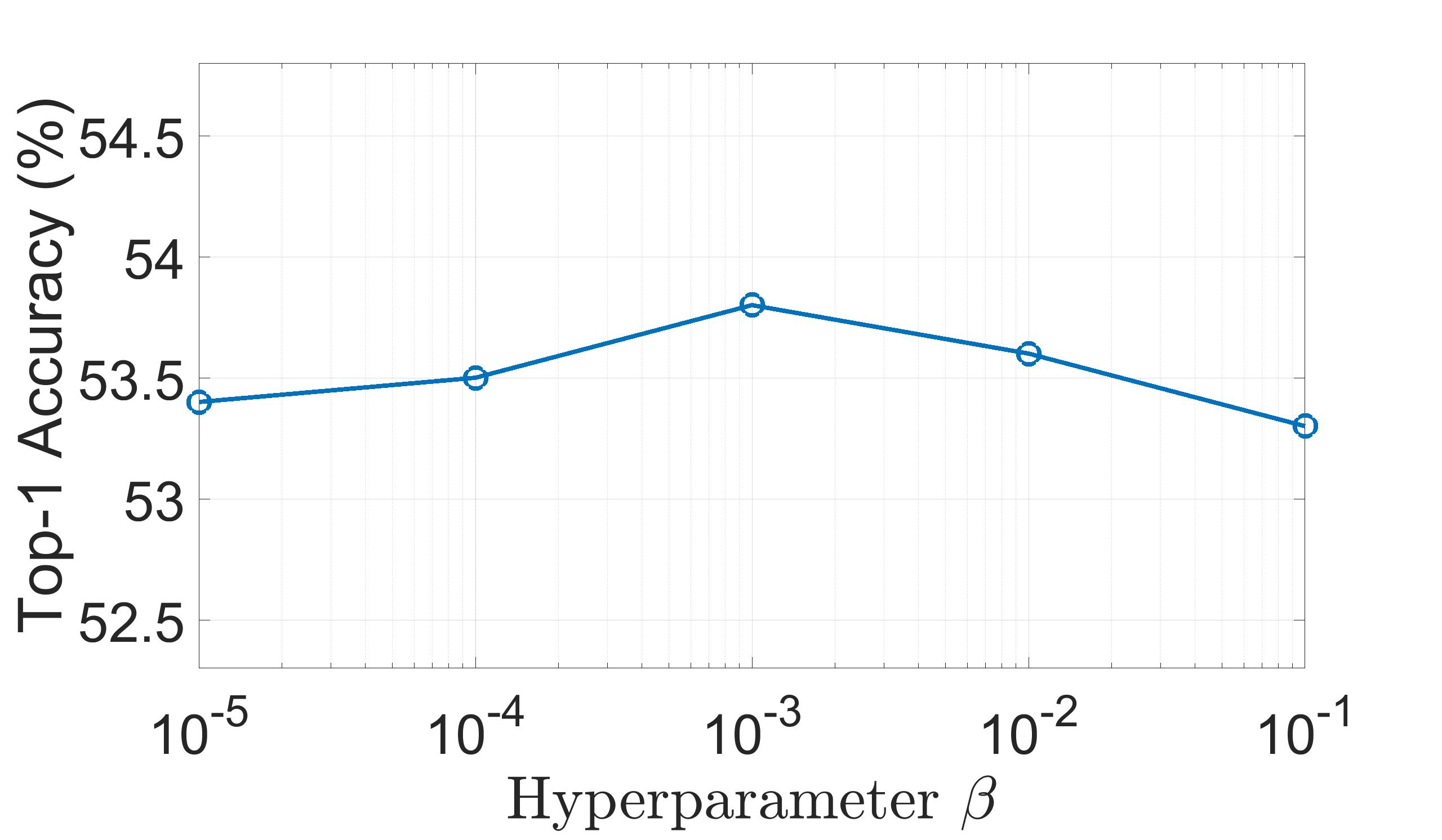}}

    \caption{Top-1 accuracy across six dataset-backbone combinations as a function of hyperparameter $\beta$. The optimal accuracy consistently emerges at $\beta = 10^{-3}$, underscoring the robustness and general adaptability of our SCS-SupCon framework. Smaller values of $\beta$ insufficiently promote style diversity, whereas excessively large values impose overly strict constraints on intra-class style variations, adversely affecting robust feature learning. Consequently, a moderate choice of $\beta=10^{-3}$ achieves optimal feature disentanglement performance.}
  \label{fig:beta_sensitivity}
\end{figure*}

\paragraph{Hyperparameter $\beta$:}
Additionally, we conducted systematic sensitivity experiments by varying the hyperparameter $\beta$ across the range $\{10^{-5},10^{-4},10^{-3},10^{-2},10^{-1}\}$, while fixing other parameters at their optimal learned values. Figure~\ref{fig:beta_sensitivity} clearly illustrates that optimal classification accuracy consistently emerges at $\beta = 10^{-3}$. Smaller values of $\beta$ are insufficient for adequately enforcing style diversity, whereas excessively large values excessively penalize intra-class style variations, adversely affecting robust feature representation. Consequently, selecting an intermediate setting ($\beta=10^{-3}$) ensures optimal performance in feature disentanglement.

Overall, the comprehensive sensitivity analyses confirm the robustness and adaptability of the proposed SCS-SupCon method, offering practical insights for effective hyperparameter selection. Particularly noteworthy are the adaptive behaviors observed, which explicitly indicate that fine-grained classification tasks notably benefit from the precise boundary adjustments provided by the sigmoid-based contrastive loss and the learnable temperature parameter. \textcolor{black}{From a practitioner’s perspective, our analysis suggests simple and robust default choices. For all datasets considered, setting the style-weight coefficient to $\beta = 10^{-3}$ yields near-optimal performance (Fig.~\ref{fig:beta_sensitivity}); values in the neighboring range $[5\times10^{-4},5\times10^{-3}]$ typically produce very similar results, which indicates that the method is not overly sensitive to moderate perturbations of this parameter.}

\textcolor{black}{We therefore adopt an initial log-temperature $t' = \log(0.1)$ (corresponding to $t = 0.1$) and $b = 0$ as default initial values across all main experiments, while both parameters remain learnable and are learned jointly with the encoder and projection head. In Section~\ref{subsec:bayesopt} we further show, through Bayesian hyperparameter optimization on five datasets, that the automatically selected $t_0^\star$ and $b_0^\star$ remain in a narrow neighborhood around these initializations, which provides additional evidence that SCS-SupCon does not require heavy manual initialization of these two parameters. The learning rates and batch sizes in Table~\ref{tab:hyperparam} follow established conventions for ResNet and ViT backbones. Crucially, our experiments show that small variations to these values do not alter the relative performance ranking of the compared methods, underscoring the robustness of our findings. }

\subsection{\textcolor{black}{Bayesian Hyperparameter Optimization of Temperature and Bias}}
\label{subsec:bayesopt}

\textcolor{black}{To complement the preceding sensitivity analysis and to assess the robustness of the proposed sigmoid-based contrastive loss under an automatic tuning scheme, we conducted a small-scale study based on Bayesian hyperparameter optimization. In particular, we use Optuna’s TPE (Tree-structured Parzen Estimator) sampler~\cite{akiba2019optuna}, which implements a Bayesian optimization strategy over continuous and discrete search spaces.}

\textcolor{black}{In this experiment we focus on the two key scalar parameters of the sigmoid-based contrastive loss, namely the initial temperature $t_0$ (parameterised as $t = \exp(t')$ with $t' = \log t_0$) and the initial bias $b_0$. We select five representative datasets from our benchmark suite (CIFAR-10, CIFAR-100, Tiny-ImageNet, CUB200-2011, and Stanford Dogs) and, for each of them, we use the ResNet-18 encoder. The objective function is defined as the top-1 validation accuracy, which we aim to maximise, after training SCS-SupCon with a shortened Stage~1 schedule, while keeping all other training settings identical to those in Table~\ref{tab:hyperparam}. At each Optuna trial, the TPE sampler proposes a candidate pair $(t_0, b_0)$ from the ranges
\[
t_0 \in [0.05, 0.2] \ \text{(log-uniform)}, \qquad b_0 \in [-0.2, 0.2] \ \text{(uniform)},
\]
and we initialise the learnable parameters as $t' = \log t_0$ and $b = b_0$. We then train Stage~1 of SCS-SupCon for a reduced schedule of 200 epochs and evaluate the top-1 accuracy on a held-out \emph{validation} split constructed from the training data and strictly disjoint from the test set used for final reporting. This validation accuracy is returned to Optuna as the objective value, and we run 30 optimization trials per dataset.}

\textcolor{black}{After the search, we select the best configuration $(t_0^\star, b_0^\star)$ for each dataset and re-train SCS-SupCon using the full training schedule employed in our main experiments. The resulting optimal hyperparameters and the corresponding test accuracies are reported in Table~\ref{tab:bayesopt}. This experiment demonstrates that the proposed sigmoid-based contrastive loss can be effectively combined with standard Bayesian hyperparameter optimization tools, and that reasonable configurations of $(t_0, b_0)$ can be obtained within a modest trial budget for different datasets.} \textcolor{black}{Across all five datasets, the optimal initial temperatures $t_0^\star$ lie in a narrow interval around 0.1, with slightly smaller values for the fine-grained datasets (CUB200-2011 and Stanford Dogs), while the optimal biases $b_0^\star$ are small and positive on generic datasets and clearly higher on fine-grained ones. This behavior is broadly consistent with the dataset-dependent convergence patterns of $t$ and $b$ observed for deeper CNN and Transformer encoders in Fig.~\ref{fig:all_curves}. The corresponding test accuracies are very close to (and consistently slightly higher than) those obtained with manually chosen initializations in Table~\ref{tab:my-table}, which indicates that SCS-SupCon can be reliably deployed either with simple default settings or with a small amount of Bayesian hyperparameter search. Note that in all cases the Bayesian optimization is used only to choose the initial values $(t_0, b_0)$; during each training run the actual learnable parameters $t'$ and $b$ are still updated deterministically by gradient descent, and no full Bayesian inference over these parameters is performed.}

\begin{table}[htbp]
\centering
\caption{\textcolor{black}{Bayesian optimization of the initial temperature $t_0$ and bias $b_0$ for SCS-SupCon using a ResNet-18 encoder. For each dataset, we report the best configuration found by Optuna (TPE sampler) together with the corresponding test accuracy obtained after full Stage~1 and Stage~2 training using these initial values. The hyperparameter search itself uses only a held-out validation split constructed from the training data; the test set remains unseen throughout the optimization process.}}
\label{tab:bayesopt}
\resizebox{0.55\textwidth}{!}{
\begin{tabular}{lccc}
\hline
\textcolor{black}{Dataset} & \textcolor{black}{Best $t_0$} & \textcolor{black}{Best $b_0$} & \textcolor{black}{Test Acc (\%)} \\
\hline
\textcolor{black}{CIFAR-10}        & \textcolor{black}{0.104} & \textcolor{black}{0.022} & \textcolor{black}{95.5} \\
\textcolor{black}{CIFAR-100}       & \textcolor{black}{0.119} & \textcolor{black}{0.038} & \textcolor{black}{75.7} \\
\textcolor{black}{Tiny-ImageNet}   & \textcolor{black}{0.122} & \textcolor{black}{0.043} & \textcolor{black}{61.0} \\
\textcolor{black}{CUB200-2011}     & \textcolor{black}{0.088} & \textcolor{black}{0.089} & \textcolor{black}{62.3} \\
\textcolor{black}{Stanford Dogs}   & \textcolor{black}{0.084} & \textcolor{black}{0.078} & \textcolor{black}{67.9} \\
\hline
\end{tabular}}
\end{table}

\subsection{Ablation Study}

Table~\ref{tab:ablation} summarizes the results of ablation experiments conducted to assess the contributions of each component within the proposed SCS-SupCon framework across multiple datasets and backbone architectures. It is clear that our proposed SCS-SupCon consistently obtains the highest accuracy among all methods evaluated. Interestingly, even the \textit{Sigmoid-only} variant (without the style-distance penalty, $\beta=0$) exhibits highly competitive performance, consistently outperforming both original CS-SupCon variants (with and without overlap). This result underscores the effectiveness of explicitly modeling pairwise relationships through our sigmoid-based contrastive loss function, which effectively addresses the negative-sample dilution problem inherent in traditional InfoNCE-based approaches.

Furthermore, introducing the explicit style-distance penalty term ($\beta$) into the sigmoid-based contrastive loss consistently provides additional performance improvements across all datasets and backbone architectures. In particular, the complete SCS-SupCon configuration (Sigmoid + $\beta$) surpasses the \textit{Sigmoid-only} variant by explicitly enforcing intra-class style diversity, thus further facilitating feature disentanglement and enhancing model robustness.

In summary, these experimental findings clearly demonstrate that the proposed SCS-SupCon framework substantially boosts classification performance by effectively alleviating the negative-sample dilution issue. Moreover, our method maintains computational efficiency and structural simplicity compared to existing contrastive learning approaches, thereby achieving consistent and significant improvements in feature discrimination capability, robustness, and generalization performance.

\begin{table*}[tbp]
\begingroup
\color{black}
\caption{Ablation analysis of the proposed SCS-SupCon loss components.}
\label{tab:ablation}
\centering
\resizebox{1.05\textwidth}{!}{
\begin{tabular}{l c c c c c}
\hline
Dataset(\%) $\uparrow$ & SupCon  & CS-SupCon  & CS-SupCon w.ov. & SCS-SupCon (Sigmoid-only, $\beta=0$)  & SCS-SupCon  \\
\hline
CIFAR10 & 95.6 & 95.4 & 95.7 & 95.7 & \textbf{95.9} \\
CIFAR100 & 75.5 & 77.6 & 78.4 & 78.3 & \textbf{79.2} \\
CUB200-2011 & 89.1 & 89.8 & 89.8 & 89.7 & \textbf{90.2} \\
S. Dogs (TinyViT 5M) & 85.5 & 86.0 & 86.2 & 86.2 & \textbf{86.6} \\
S. Dogs (ConvNeXt-tiny) & 92.8 & 92.8 & 93.1 & 93.1 & \textbf{93.8} \\
Pascal VOC & 51.9 & 51.4 & 53.3 & 53.2 & \textbf{53.8} \\
\hline
\end{tabular}}
\endgroup
\end{table*}

\subsection{Statistical Significance Analysis}

To rigorously assess the statistical significance of performance differences between our proposed SCS-SupCon framework and baseline methods, we conducted the Friedman test~\cite{friedman1937use} followed by the Nemenyi post-hoc test~\cite{demvsar2006statistical}.
 The statistical significance analysis encompassed six representative dataset-encoder combinations: CIFAR-10 (ResNet-50), CIFAR-100 (ResNet-50), CUB200-2011 (TinyViT-21M), Stanford Dogs (TinyViT-5M and ConvNeXt-tiny), and PASCAL VOC 2005 (TinyViT-5M). Rankings of methods used for the statistical tests were based on classification accuracies reported in Table~\ref{tab:results}.

Specifically, given $K=15$ methods and $N=6$ datasets, the Friedman statistic is computed as:
\begin{equation}
\chi_F^2 = \frac{12N}{K(K+1)}\left[\sum_{j=1}^K R_j^2 - \frac{K(K+1)^2}{4}\right],
\end{equation}
where $R_j$ denotes the average ranking of the $j$-th method across all datasets.

After confirming statistical significance with the Friedman test (at significance level $\alpha = 0.05$), we further performed the Nemenyi post-hoc test to pinpoint pairwise differences among methods. The critical difference (CD) is calculated as:
\begin{equation}
CD = q_\alpha \sqrt{\frac{K(K+1)}{6N}},
\end{equation}
where $q_\alpha$ represents the critical value derived from the Studentized range distribution.

\begin{figure}[htbp] 
\centering 
\includegraphics[width=0.9\linewidth]{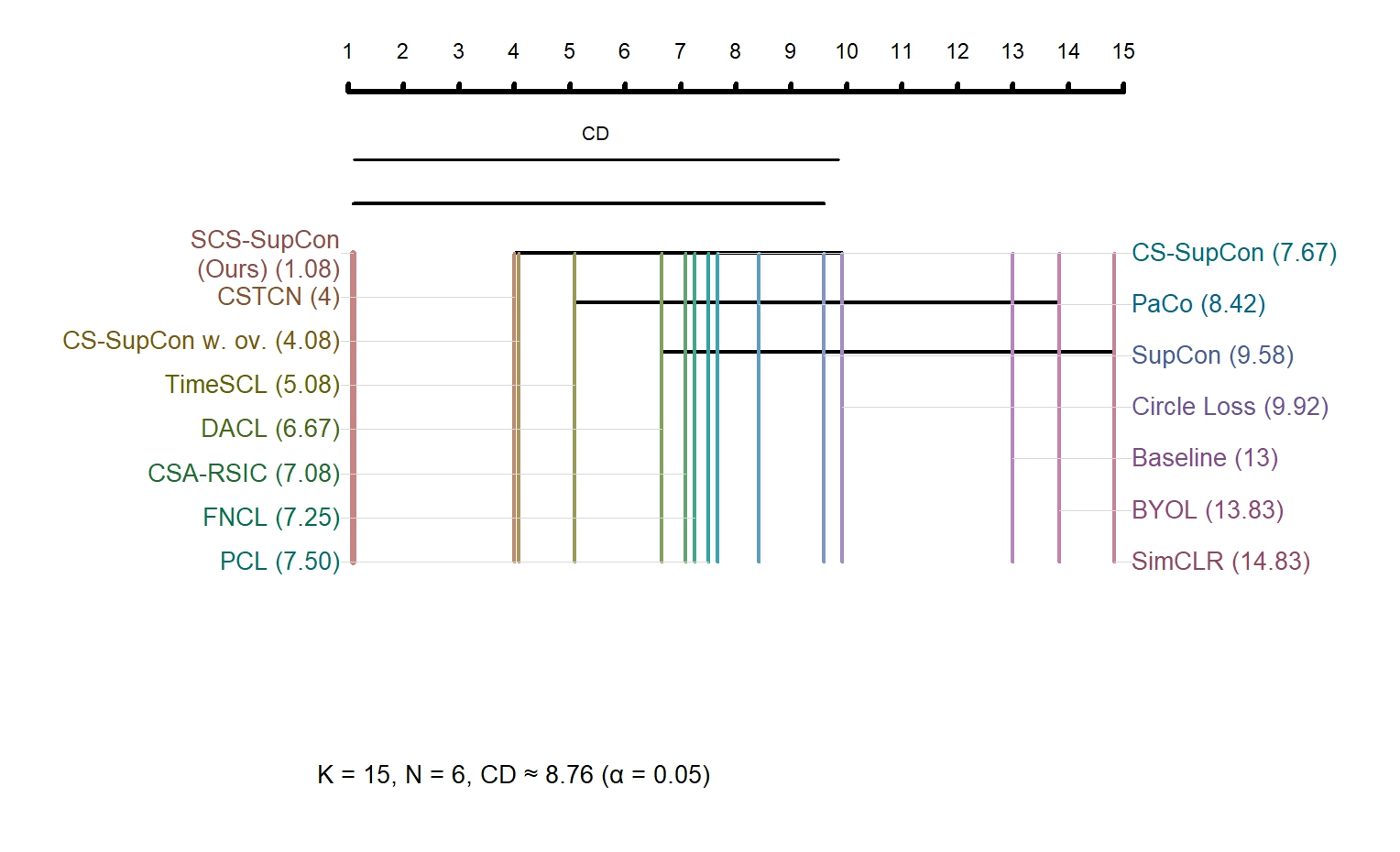} 
\caption{Critical Difference (CD) diagram of the Nemenyi post-hoc test for 15 methods across six datasets (CD = 8.76 at 0.05 significance level). Methods whose average ranks differ by less than the CD value are connected by horizontal lines, indicating no statistically significant performance differences. Different colors and line thicknesses are employed to enhance visual clarity and distinguish methods clearly.} 
\label{fig:cd_diagram} 
\end{figure}

The Friedman test results ($\chi^2_F$, $p < 0.05$) demonstrated significant differences among the evaluated methods, clearly rejecting the null hypothesis that all methods exhibit equal performance. Subsequently, the Nemenyi test results (illustrated through the critical difference diagram in Figure~\ref{fig:cd_diagram}) indicate that our proposed SCS-SupCon method achieves the highest average ranking among all evaluated methods.

\textcolor{black}{In our experimental setting with $K=15$ methods and $N=6$ datasets, the standard Nemenyi formula yields a critical difference of $CD = q_{0.05}\sqrt{\tfrac{K(K+1)}{6N}} \approx 8.76$ when using the two-sided Studentised range statistic $q_{0.05}$ tabulated in~\cite{demvsar2006statistical}. This relatively large value of $CD$ reflects the combination of a small number of datasets and a comparatively large number of methods, which inflates the standard error of the average ranks and therefore leads to a conservative threshold for detecting significant differences. In such a regime, only very pronounced rank gaps can be declared statistically significant, even though smaller yet consistent improvements—such as those obtained by SCS-SupCon across the considered datasets—may still be practically relevant.}

In the CD diagram (Figure~\ref{fig:cd_diagram}), the best-performing method corresponds to rank 1 (leftmost position), while methods with higher numerical ranks represent comparatively worse performance. Although the statistical significance analysis indicates that performance differences among top-ranking methods remain subtle and do not reach strict statistical significance, our proposed SCS-SupCon consistently exhibits robust and sustained superior performance across all evaluated datasets and encoder architectures, achieving the best overall average ranking. These results clearly demonstrate the competitiveness and reliability of our method within the highly challenging contrastive learning domain.

\section{\textcolor{black}{Discussion}}
\label{sec6}

\textcolor{black}{This section discusses the broader implications of the proposed SCS-SupCon framework, analyses its strengths and limitations, and provides practical guidance for future applications.}

\textcolor{black}{\textbf{Why sigmoid-based contrastive loss adaptive boundaries help in fine-grained settings.}
Fine-grained recognition problems are characterized by small inter-class margins and substantial intra-class variability. In such regimes, InfoNCE-based supervised contrastive losses suffer from diluted gradients due to a large number of easy negatives and fixed decision boundaries constrained by the temperature parameter. By employing a sigmoid-based pairwise contrastive loss with learnable decision-boundary parameters $(t', b)$, SCS-SupCon adaptively concentrates gradients around ambiguous pairs and performs a pairwise contrastive learning, significantly benefiting tasks with subtle distinctions between classes, such as CUB200-2011 and Stanford Dogs.}

\textcolor{black}{\textbf{Practical significance of performance gains.}
Although improvements are modest on saturated benchmarks such as CIFAR-10, our proposed SCS-SupCon method yields more pronounced accuracy gains on challenging fine-grained classification tasks (see Tables~\ref{tab:results} and~\ref{tab:my-table}), with negligible computational overhead (Table~\ref{tab:efficiency}). Notably, our method builds directly upon the original non-overlapping CS-SupCon formulation, offering a simpler and more interpretable approach compared to the overlapping variant, which is less intuitive and harder to extend in future research. Such performance enhancements are practically meaningful, particularly in fields like medical imaging and industrial inspection, where even modest accuracy improvements can substantially reduce errors without architectural complexity or additional computational cost.}

\textcolor{black}{\textbf{Guidelines and limitations of hyperparameter tuning.}
Our experiments confirm that SCS-SupCon is somewhat more sensitive to the choice of its hyperparameter $\beta$ (and to the initialization of the learnable boundary parameters $t'$ and $b$) than simpler InfoNCE-based baselines, which is a common characteristic of contrastive learning methods. Nevertheless, the analyses in Section~\ref{sec5} indicate that a small set of default values works reliably across diverse datasets: we set $\beta = 10^{-3}$, initialise $t$ to $0.1$ (with $t' = \log 0.1$), and initialise $b$ to $0$ as good starting points. In practice, we recommend fixing the initial values of $t$ and $b$ to these initializations and conducting a coarse search over $\beta$ in $[5\times 10^{-4}, 5\times 10^{-3}]$ if computational resources permit. Beyond the small-scale Bayesian optimization study presented in Section~\ref{subsec:bayesopt}, a more sophisticated avenue for future work is to couple SCS-SupCon with automated hyperparameter optimization or meta-learning strategies that jointly tune a larger set of training hyperparameters. Such techniques could further reduce the need for manual tuning and improve robustness in large-scale deployments.}

\textcolor{black}{\textbf{Failure cases and scope of applicability.}
Although SCS-SupCon consistently matches or outperforms strong baselines across all considered benchmarks, the relative advantage over CS-SupCon with overlap is small on some easier datasets (e.g., CIFAR-10). This observation suggests that when the underlying task is relatively coarse-grained or already well separated by the backbone, the additional flexibility provided by adaptive boundaries may yield diminishing returns. Furthermore, SCS-SupCon inherits the typical computational characteristics of pairwise contrastive objectives: its per-iteration cost scales quadratically with the batch size, which may limit its direct applicability to extremely large batch regimes unless combined with sampling or memory-bank techniques. We therefore view SCS-SupCon as particularly well suited to scenarios where fine-grained discrimination and robustness to intra-class variation are critical, while acknowledging that simpler losses may remain competitive for very low-resolution or easily separable tasks.}

\textcolor{black}{\textbf{Interpretation of statistical significance results.}
Regarding the statistical analysis in Section~\ref{sec5}, we apply the Friedman and Nemenyi tests to compare all methods. With $K=15$ methods and $N=6$ datasets, the critical difference (CD) at the 0.05 significance level evaluates to $CD = 8.76$, as reported in Figure~\ref{fig:cd_diagram}. This relatively large CD is a direct consequence of the combination of a limited number of datasets and a comparatively large number of competing methods: in such a regime, the standard error of the average rank estimates is non-negligible, and only very large differences in rank can be declared statistically significant. Consequently, the fact that the top methods are connected in the CD diagram does not imply that they are indistinguishable in practice; rather, it reflects the conservative nature of the Nemenyi test under scarce sample sizes. Importantly, SCS-SupCon attains the best average rank among all methods and exhibits consistent improvements across all six datasets (Tables~\ref{tab:results} and~\ref{tab:my-table}), which we consider strong evidence of its practical advantage despite the strictness of the multiple-comparison correction.}

\textcolor{black}{\textbf{Future extensions.}
Finally, our framework opens several promising directions for future research. First, combining SCS-SupCon with large-scale pre-trained vision transformers and conducting experiments on ultra-large and cross-domain benchmarks (e.g., ImageNet and long-tailed or domain-shifted variants) would provide a more comprehensive assessment of its scalability. Second, the explicit separation between common and style features makes SCS-SupCon a natural candidate for semi-supervised or open-set extensions, where pseudo-labelling or self-supervised objectives could be attached to the style field while keeping the common field supervised. Third, the interpretability of the learned boundaries could be further improved by linking the learned $(t',b)$ parameters to task-specific notions of similarity or by integrating fairness-aware regularisers when deploying the model in sensitive application domains.}

\textcolor{black}{\;In addition, we have incorporated recent contrastive advances, including diffusion-based contrastive and hierarchical transformer-based SupCon variants, into our Related Work and view them as complementary directions; a thorough empirical comparison with these methods, especially on larger and more diverse benchmarks, is an interesting avenue for future research.}

\textcolor{black}{\textbf{Ethical and practical considerations.}
As with any advanced learning framework, deploying SCS-SupCon in real-world applications—especially in safety-critical domains such as medical diagnosis or surveillance—requires attention to fairness, transparency, and robustness. Because our method can amplify subtle patterns in the data, it is important to ensure that the training set is representative of the target population and that potential biases are monitored and mitigated. We therefore recommend combining SCS-SupCon with established practices for responsible AI, including bias analysis, calibration checks, and, where possible, post-hoc interpretability techniques to help practitioners understand which feature dimensions drive the learned boundaries.}

\section{Conclusion}
\label{sec7}

This paper introduces SCS-SupCon, a novel supervised contrastive learning framework specifically designed to address the negative-sample dilution problem and the lack of explicit decision boundary mechanisms prevalent in existing contrastive learning methods. By proposing a sigmoid-based pairwise contrastive loss with adaptive decision boundaries parameterized by learnable temperature and bias terms, SCS-SupCon effectively emphasizes discriminative information from hard negative samples. Additionally, the explicit integration of a style-distance constraint enables robust disentanglement between category-relevant (common) and category-irrelevant (style) features. Comprehensive experimental evaluations conducted on six benchmark datasets demonstrate that our proposed approach consistently achieves state-of-the-art performance, surpassing classical methods (SupCon), recent supervised methods (SelfCon, CS-SupCon, TimeSCL, FNCL), clustering-based approaches (PCL, CSTCN, Circle Loss), and hybrid or adversarial methods (PaCo, DACL, CSA-RSIC). Rigorous ablation studies and statistical significance analyses further confirm the effectiveness, robustness, and generalization capabilities of our proposed framework.

The key advantages of SCS-SupCon include significantly enhanced discriminative power, particularly in fine-grained classification tasks, and effective adaptive decision boundary control facilitated by learnable parameters. Nevertheless, although SCS-SupCon already operates on cosine-based similarities, it does not explicitly enforce angular margins between classes, so there is still room to further enhance the angular discriminative power of the learned representations in extremely challenging fine-grained scenarios.

Given that explicit angular discrimination remains a central challenge in metric learning and fine-grained recognition, our future work will focus on developing more angular-discriminative variants of SCS-SupCon, for example by combining its adaptive decision boundaries with explicit angular margins or other cosine-based objectives. Additionally, we intend to explore lightweight training strategies and extend SCS-SupCon to semi-supervised and incremental learning contexts, thereby broadening its practicality and effectiveness in diverse real-world applications.


\end{document}